\documentclass{article}

\usepackage{microtype}
\usepackage{graphicx}
\usepackage{subcaption}
\usepackage{booktabs} % for professional tables

\usepackage{hyperref}

\usepackage[preprint]{icml2026}

\usepackage{amsmath}
\usepackage{amssymb}
\usepackage{mathtools}
\usepackage{amsthm}

\usepackage[capitalize,noabbrev]{cleveref}

\theoremstyle{plain}

\theoremstyle{definition}

\theoremstyle{remark}

\usepackage[textsize=tiny]{todonotes}
\usepackage{url}
\usepackage{amsfonts}
\usepackage{nicefrac}
\usepackage{dsfont}
\usepackage{wrapfig}
\usepackage{array}
\usepackage{bbding}
\usepackage{pifont}
\usepackage{bm}
\usepackage{float}
\usepackage{makecell}
\usepackage{enumitem}
\usepackage{multirow}
\usepackage{tcolorbox}
\usepackage{xcolor}
\usepackage{listings}
\usepackage{fancyvrb}

\newtcolorbox{coloredtextbox}[2][]{
    colback=#2!10,             % 主体背景颜色，带透明度
    colframe=#2,               % 边框颜色
    coltitle=white,            % 标题文字颜色
    colbacktitle=#2,           % 标题框背景颜色
    title=#1,                  % 标题内容
    fonttitle=\bfseries,       % 标题字体样式
    sharp corners=south,       % 主体框圆角，下方圆角
    arc=2mm,                   % 主体框圆角半径
    boxrule=1mm,               % 边框粗细
    width=0.5\textwidth,          % 文本框宽度
}

\newtcolorbox{coloredtextbox1}[2][]{
    colback=#2!10,             % 主体背景颜色，带透明度
    colframe=#2,               % 边框颜色
    coltitle=white,            % 标题文字颜色
    colbacktitle=#2,           % 标题框背景颜色
    title=#1,                  % 标题内容
    fonttitle=\bfseries,       % 标题字体样式
    sharp corners=south,       % 主体框圆角，下方圆角
    arc=2mm,                   % 主体框圆角半径
    boxrule=1mm,               % 边框粗细
    width=\textwidth,          % 文本框宽度
}

\def\approach{ReLayout}

\icmltitlerunning{\approach{}: Versatile and Structure-Preserving Design Layout Editing via Relation-Aware Design Reconstruction}

\begin{document}

\twocolumn[
  \icmltitle{\approach{}: Versatile and Structure-Preserving Design Layout Editing \\ via Relation-Aware Design Reconstruction}

  % It is OKAY to include author information, even for blind submissions: the
  % style file will automatically remove it for you unless you've provided
  % the [accepted] option to the icml2026 package.

  % List of affiliations: The first argument should be a (short) identifier you
  % will use later to specify author affiliations Academic affiliations
  % should list Department, University, City, Region, Country Industry
  % affiliations should list Company, City, Region, Country

  % You can specify symbols, otherwise they are numbered in order. Ideally, you
  % should not use this facility. Affiliations will be numbered in order of
  % appearance and this is the preferred way.
  \icmlsetsymbol{equal}{*}

  \begin{icmlauthorlist}
    \icmlauthor{Jiawei Lin}{xjtu}
    \icmlauthor{Shizhao Sun}{microsoft}
    \icmlauthor{Danqing Huang}{microsoft}
    \icmlauthor{Ting Liu}{xjtu}
    \icmlauthor{Ji Li}{microsoft}
    \icmlauthor{Jiang Bian}{microsoft}
    %\icmlauthor{}{sch}
    %\icmlauthor{}{sch}
  \end{icmlauthorlist}

  \icmlaffiliation{xjtu}{Xi'an Jiaotong University, Xi'an, China}
  \icmlaffiliation{microsoft}{Microsoft Research}

  \icmlcorrespondingauthor{Jiawei Lin}{kylelin@stu.xjtu.edu.cn}
  \icmlcorrespondingauthor{Shizhao Sun}{shizsu@microsoft.com}

  % You may provide any keywords that you find helpful for describing your
  % paper; these are used to populate the "keywords" metadata in the PDF but
  % will not be shown in the document
  % \icmlkeywords{Machine Learning, ICML}

  \vskip 0.3in
]

% this must go after the closing bracket ] following \twocolumn[ ...

% This command actually creates the footnote in the first column listing the
% affiliations and the copyright notice. The command takes one argument, which
% is text to display at the start of the footnote. The \icmlEqualContribution
% command is standard text for equal contribution. Remove it (just {}) if you
% do not need this facility.

% Use ONE of the following lines. DO NOT remove the command.
% If you have no special notice, KEEP empty braces:
\printAffiliationsAndNotice{}  % no special notice (required even if empty)
% Or, if applicable, use the standard equal contribution text:
% \printAffiliationsAndNotice{\icmlEqualContribution}

\begin{abstract}
Automated redesign without manual adjustments marks a key step forward in the design workflow.
In this work, we focus on a foundational redesign task termed design layout editing, which seeks to autonomously modify the geometric composition of a design based on user intents.
To overcome the ambiguity of user needs expressed in natural language, we introduce four basic and important editing actions and standardize the format of editing operations.
The underexplored task presents a unique challenge: satisfying specified editing operations while simultaneously preserving the layout structure of unedited elements.
Besides, the scarcity of triplet (original design, editing operation, edited design) samples poses another formidable challenge.
To this end, we present \approach{}, a novel framework for versatile and structure-preserving design layout editing that operates without triplet data.
Specifically, \approach{} first introduces the relation graph, which contains the position and size relationships among unedited elements, as the constraint for layout structure preservation.
Then, relation-aware design reconstruction (RADR) is proposed to bypass the data challenge.
By learning to reconstruct a design from its elements, a relation graph, and a synthesized editing operation, RADR effectively emulates the editing process in a self-supervised manner. 
A multi-modal large language model serves as the backbone for RADR, unifying multiple editing actions within a single model and thus achieving versatile editing after fine-tuning.
Qualitative, quantitative results and user studies show that \approach{} significantly outperforms the baseline models in terms of editing quality, accuracy, and layout structure preservation.
\end{abstract}
\section{Introduction}
\label{sec:intro}

\begin{figure*}
    \centering
    \includegraphics[width=\linewidth]{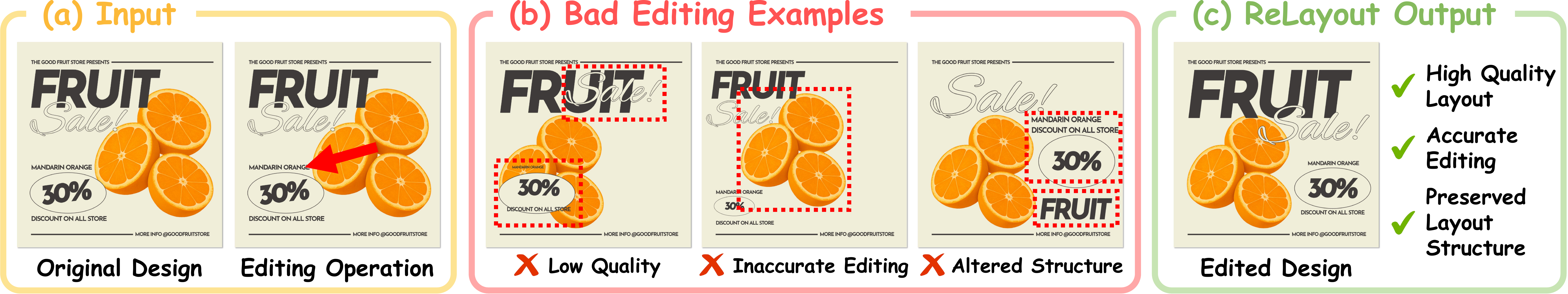}
    \caption{(a) Design layout editing receives an original design and an editing operation (e.g., move) as input. The red arrow indicates the moving element (orange image) and the target position. (b) Design layout editing is a challenging task. We visualize some typical undesirable results, including the low-quality layout, inaccurate editing, and destroyed layout structure. These issues are marked with red dotted boxes. (c) The proposed \approach{} enables effective design layout editing, as evidenced by the high layout quality, accurate editing, and well-preserved layout structure in the edited design.}
    \label{fig:teaser}
\end{figure*}

Graphic design is an iterative process that involves multiple redesigns to explore variations or update design preferences.
Among various redesigns, design layout editing, which focuses on adjusting the layout of an input design, plays a fundamental yet crucial role.
For instance, one might enlarge the text box to emphasize a key message, or reposition a product image for better visual appeal (Figure \ref{fig:teaser}a). 
During design layout editing, the geometric composition of the input design is updated according to user requirements.

Performing such edits, however, includes multi-dimensional considerations and sets a high barrier for users.
The edited design should have good visual quality and meet the specified editing operation.
In addition, the layout structure among unedited elements should be preserved as much as possible to maintain the original design intent, reading order, etc.
Figure~\ref{fig:teaser}b shows some common undesirable results.
When repositioning the product image, if surrounding elements are not adjusted accordingly, occlusion issues may arise, thereby degrading the overall quality.
For the unedited heading ``FRUIT", which is positioned at the top of the canvas in the original design, it is preferable not to move it to the bottom in the edited design.

In this work, we aim to automate the design layout editing process, paving the way for a more efficient and accessible design workflow.
Given an original design and an element-level editing operation (Figure~\ref{fig:teaser}a), the target output is an edited design (Figure~\ref{fig:teaser}c).
To improve clarity and disambiguation, we do not use natural language for editing operations.
Instead, we use a standardized format to clearly specify the editing action, the target element, and corresponding parameters (e.g., move element 1 to (100, 100)).
For editing actions, we consider four of the most practical ones: add, delete, move, and resize.
Please note that we exclude content-related actions (e.g., change the object within a product image), as they fall within the scope of image editing rather than layout editing.

Despite its importance, automatic design layout editing has received limited attention.
Currently, the most relevant work is layout generation~\cite{jyothi2019layoutvae, kong2022blt, li2020attribute, Kikuchi2021, jiang2023layoutformerconditionalgraphiclayout, inoue2023layoutdmdiscretediffusionmodel, lin2023parse} and design generation~\cite{inoue2023document, cheng2024graphic, Shabani_2024_CVPR, lin2024elements}.
They represent the layout/design as a set of element attributes (e.g., x, y, width, height, color, font, etc) and explore various conditions for controllable generation.
However, none of them can take an existing design as the input condition to produce an edited version.
Another line of related work lies in image editing methods~\cite{brooks2023instructpix2pix, kawar2023imagic, shi2024seededit, zhang2023magicbrush, zhang2025context}, which have made remarkable progress with the advancement of text-to-image diffusion models~\cite{podell2023sdxl, esser2024scaling, peebles2023scalable, flux2024}.
Nevertheless, directly applying these pixel-level editing models to design layout editing is problematic because they inevitably alter content details.
Furthermore, pixel-space output hinders further editing.

Design layout editing faces unique challenges.
First, it requires preserving the layout structure of unedited elements, i.e., their relative size and position relationships.
Such relationships, which convey the design’s intended reading order, visual focus, and hierarchy, should be retained as much as possible after editing.
Second, the lack of editing data (original design, editing operation, edited design) fundamentally hinders the development of this task.
Although some datasets~\cite{deka2017rico, yamaguchi2021canvasvae} provide high-quality graphic designs, they don't contain such triplets.
Also, manually constructing such samples is prohibitively expensive and difficult to scale.

To this end, we propose \textit{\approach{}}, a novel framework designed to tackle the above challenges and achieve design layout editing effectively.
First, we employ the relation graph in \approach{} as an explicit editing constraint.
By representing all unedited elements and the canvas as nodes, while extracting their size and position relationships as edges, the relation graph faithfully captures the original layout structure and can work as an additional input to control the editing process, ensuring layout structure preservation.
Based on the relation graph, we present relation-aware design reconstruction (RADR) to address data scarcity.
Our key insight is that editing knowledge can be learned from individual designs via a self-supervised objective, thereby bypassing the need for triplet training data.
Specifically, RADR proposes to reconstruct a design from its elements, a relation graph, and a randomly synthesized editing operation, which mirrors the objective of design layout editing and enables direct generalization to the target task.
Following the successful practices of prior approaches~\cite{lin2024elements, cheng2024graphic}, we leverage a multi-modal large language model as the backbone of RADR and fine-tune it to deal with multiple editing actions.
This eventually yields a versatile and structure-preserving design layout editing model in the absence of explicit editing dataset.

We evaluate the performance from different perspectives: editing quality, editing accuracy, and layout structure preservation.
Since \approach{} is the first approach for this task, we establish a baseline using GPT-4o~\cite{hurst2024gpt}, and also compare \approach{} with state-of-the-art design generation models in terms of output quality.
Qualitative results, quantitative results and user studies show that \approach{} outperforms the baselines in all evaluation dimensions.
Furthermore, we extend \approach{} to achieve other practical tasks, including language-guided design layout editing and composite design layout editing.
\section{Related Work}
\label{sec:related work}

\begin{figure*}[t]
    \centering
    \includegraphics[width=\linewidth]{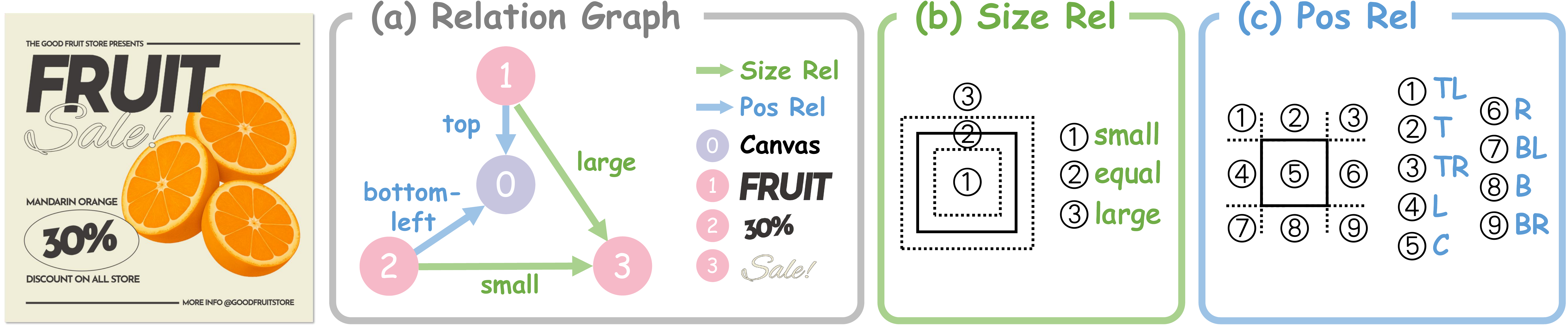}
    \caption{
    (a) An example of the relation graph. For simplicity, only a subset of nodes and edges are visualized. The node at the non-arrow end of each edge represents the source, while the node at the arrow end represents the target.
    (b) and (c) Heuristic rules for detecting size and position relationships. The solid-line box denotes the target node. L, R, T, B, and C stand for left, right, top, bottom, and center.}
    \label{fig:relationship}
\end{figure*}

\noindent \textbf{Layout Generation.}
Automatic layout generation is an emerging topic in computer vision.
Existing methods develop various layout generation models, covering unconditional layout generation~\cite{Arroyo_2021_CVPR, li2020layoutgan, jiang2022coarse} and conditional layout generation from the element type~\cite{jyothi2019layoutvae, kong2022blt, li2020attribute}, size~\cite{kong2022blt}, relationship~\cite{Kikuchi2021, lee2020neuraldesignnetworkgraphic}, or textual description~\cite{lin2023parse}.
Some recent studies further unify multiple generation tasks within a single model~\cite{jiang2023layoutformerconditionalgraphiclayout, tang2023layoutnuwarevealinghiddenlayout, inoue2023layoutdmdiscretediffusionmodel}. 
However, none of these approaches investigate the layout editing task.
To be best of our knowledge, \approach{} is the first framework for design layout editing, which effectively reduces the workload during the redesign process.

\noindent \textbf{Design Generation.}
Compared to layout generation models that merely produce layouts (i.e., element position and size), design generation models create the whole design.
To achieve this, FlexDM~\cite{inoue2023document} represents each element as a set of multi-modal fields and employs a masked field prediction strategy to fill the masked element attributes.
VLC~\cite{Shabani_2024_CVPR} combines an image diffusion model with a vector diffusion model for design generation, taking the advantages of both domains to enhance quality.
Graphist~\cite{cheng2024graphic} formulates the task as a sequence-to-sequence problem.
It fine-tunes a multi-modal large language model (MLLM) to predict JSON-formatted attribute sequences based on the element content sequences.
LaDeCo~\cite{lin2024elements} further improves generation quality by introducing the layered design principle.
While these approaches show promise, they remain incapable of performing design layout editing.

\noindent \textbf{Editing in Other Domains.}
Editing is a crucial task in various content creation domains, e.g., images~\cite{brooks2023instructpix2pix, kawar2023imagic, shi2024seededit, zhang2023magicbrush, zhang2025context}, indoor scenes~\cite{zheng2024editroomllmparameterizedgraphdiffusion, yang2024llplace3dindoorscene}, and 3D objects~\cite{Haque_2023_ICCV, chen2023gaussianeditor,yuan2025cadeditorlocatetheninfillframeworkautomated}.
In contrast to them, design layout editing poses unique challenges.
Specifically, editing an element often triggers adjustments of other unedited elements, yet their layout structure should be preserved as much as possible.
While editing tasks in other domains are more localized, such as adding a mustache to a 3D face~\cite{Haque_2023_ICCV}.

\section{Task Definition}
\label{sec:task}

\begin{figure*}
    \centering
    \includegraphics[width=\linewidth]{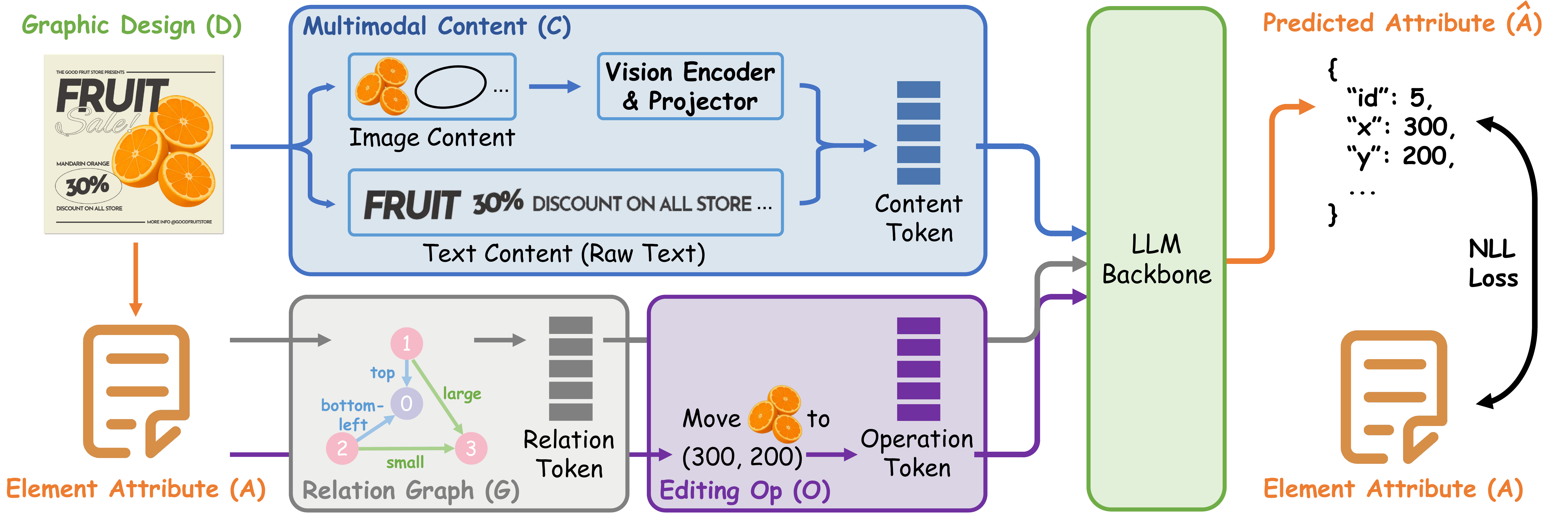}
    \caption{Illustration of RADR, which achieves design layout editing via a self-supervised objective: reconstructing a design $\mathcal{D}$($\mathcal{A}$) from its elements $\mathcal{C}$, the relation graph $\mathcal{G}$, and an editing operation $\mathcal{O}$. An MLLM serves as the backbone, using the NLL loss on attributes $\mathcal{A}$.}
    \label{fig:model}
\end{figure*}

Let $\mathcal{D}$ denote a graphic design and $\mathcal{O}$ denote an editing operation.
Design layout editing can be formally written as a function $f: (\mathcal{D}_{\text{in}}, \mathcal{O}) \mapsto \mathcal{D}_{\text{out}}$, which takes an original design $\mathcal{D}_{\text{in}}$ and an editing operation $\mathcal{O}$ as input, and produces an edited design $\mathcal{D}_{\text{out}}$.
$f$ should meet several key requirements: (1) preserving the layout structure of unedited elements in $\mathcal{D}_{\text{in}}$, (2) accurately performing the specified editing operation $\mathcal{O}$, and (3) outputting a high-quality edited design $\mathcal{D}_{\text{out}}$ accordingly.
To clarify this task, in what follows, we elaborate on $\mathcal{D}$ and $\mathcal{O}$ to explain how graphic designs are structured and how editing operations are specified.

For a design $\mathcal{D}$, it consists of multiple individual elements $\{E_i\}_{i=1}^N$ (e.g., title, icon), where $N$ is the number of elements and $i$ is the element index.
Each element has its own content and attribute set, denoted as $C_i$ and $A_i$ for the $i$-th element, respectively.
In this work, we consider two element modalities: image and text.
For the image modality, the content is RGB pixel values, and the attributes are four coordinates used to locate its position and size, i.e., center-x, center-y, width, and height.
For the text modality, the content is raw text (e.g., ``Happy birthday''), with additional attributes such as font, color, text align, etc.

We standardize the format of $\mathcal{O}$ to include three components: an editing action, the target element of that action, and any required parameters.
Four of the most practical actions are considered in this work: move, resize, add, and delete.
For the move action, the new center-x and center-y coordinates should be specified as the parameters.
For the resize action, the parameters are the new width and height.
The add and delete actions do not have additional parameters.

\section{\approach{}}
\label{sec:approach}

\subsection{Overview}
\label{subsec:overview}

Let $\mathcal{C} := \{C_i\}_{i=1}^N$ and $\mathcal{A} := \{A_i\}_{i=1}^N$, where $C_i$ and $A_i$ have been defined in Section~\ref{sec:task}.
We have $\mathcal{D}_\text{in} = (\mathcal{C}, \mathcal{A}_\text{in})$, $\mathcal{D}_\text{out} = (\mathcal{C}, \mathcal{A}_\text{out})$, where $\mathcal{A}_\text{in}$ and $\mathcal{A}_\text{out}$ are complete element attributes of the input and edited designs.
$\mathcal{C}$ is the set containing all element content, and it is the same in both designs.
Hence, we can rewrite the function $f$ in Section~\ref{sec:task} as $f: (\mathcal{C}, \mathcal{A}_\text{in}, \mathcal{O}) \mapsto \mathcal{A}_{\text{out}}$.
However, the lack of available datasets makes it infeasible to train such a mapping.
In \approach{}, we first construct a relation graph $\mathcal{G}$ from $\mathcal{A}_\text{in}$ to capture the layout structure of unedited elements in $\mathcal{D}_{\text{in}}$ (Section~\ref{subsec:graph}).
Then, relation-aware design reconstruction (RADR) is proposed, which takes $\mathcal{G}$ as an explicit constraint to control the editing process.
We implement RADR using a multi-modal large language model (MLLM), fine-tuning it to handle four actions simultaneously (Section~\ref{subsec:self-supervise}).
This yields a versatile and structure-preserving design layout editing model.

\subsection{Relation Graph}
\label{subsec:graph}

Let $\mathcal{G} := (\mathcal{V}, \mathcal{R})$ be the relation graph of a design $\mathcal{D}$, where $\mathcal{V}$ represents nodes and $\mathcal{R}$ represents edges.
Its nodes contain all elements in $\mathcal{D}$.
Since the canvas $E_0$ plays a crucial role in defining the layout structure, we add it as an extra node, i.e., $\mathcal{V} = \{E_i\}_{i=0}^{N}$ (Figure~\ref{fig:relationship}(a)).
Each edge $R_{ij} \in \mathcal{R}$ captures the position or size relationship between a source node $E_i$ and a target node $E_j$.
We define three types of size relationships (Figure~\ref{fig:relationship}(b)) and nine types of position relationships (Figure~\ref{fig:relationship}(c)).
Both are identified from $\mathcal{A}_\text{in}$ using heuristic rules.

\noindent \textbf{Size Relationship.}
For a source node $E_i$ and a target node $E_j$ in $\mathcal{G}$, their size relationship $R_{ij}^s$ is determined based on the area ratio (AR) (Figure~\ref{fig:relationship}(b)): $\text{AR} = w_ih_i/w_jh_j$,
where $w$ and $h$ denote the width and height, respectively.
The size relationship is classified as follows:
(1) $R_{ij}^s = \texttt{small}$ if $\text{AR} < 1 - \alpha$,
(2) $R_{ij}^s = \texttt{equal}$ if $1 - \alpha \leq \text{AR} \leq 1 + \alpha$,
(3) $R_{ij}^s = \texttt{large}$ if $\text{AR} > 1 + \alpha$.
Here, $\alpha$ is a predefined tolerance factor.
Notably, no size relationship is established between the canvas node and other element nodes, as the canvas is typically much larger than individual elements, making such relationships unnecessary.

\noindent \textbf{Position Relationship.}
There are two cases for position relationships $R_{ij}^p$.
First, when both the source and target nodes are elements, their position relationship can be defined using a $3 \times 3$ grid partitioning based on the target node's bounding box.
Specifically, we extend along the four edges of the bounding box, dividing the surrounding space into nine regions.
The position relationship is then assigned according to the region where the center point of the source node falls (Figure~\ref{fig:relationship}(c)).
Second, when the source node is an element and the target node is the canvas, we employ a $3 \times 3$ equal grid division \textit{within} the canvas to determine the element’s position relative to it.
Notably, no edges originate from the canvas in the position relationship.
This is because the canvas serves as a reference for positioning other elements and is only used as a target node.

Since the relationships between element nodes are bidirectional (i.e., edges 1→2 and 2→1), we randomly select one as the valid edge to avoid information redundancy.

\begin{table*}
    \centering
    \resizebox{0.9\textwidth}{!}{
    \begin{tabular}{lcccccccccc}
    \toprule
        \multirow{2}{*}{Methods} & \multicolumn{5}{c}{LLaVA-OV Scores $\uparrow$} & \multirow{2}{*}{Ove $\downarrow$} & \multirow{2}{*}{Ali $\downarrow$} & \multirow{2}{*}{Size Rel $\uparrow$} & \multirow{2}{*}{Pos Rel $\uparrow$} & \multirow{2}{*}{Op $\uparrow$} \\
        % & Layout & Image & Typography & Content & Innovation & & & & & & & & \\
        & (i) & (ii) & (iii) & (iv) & (v) & & & & & \\
        \midrule
        % PosterLLaVa~\cite{yang2024posterllava} & - & - & - & - & - & 0.1088 & 0.0015 & 0.8388 & 0.8255 & - \\
        % \midrule
        % FlexDM~\cite{inoue2023document} & 5.34 & 5.29 & 5.41 & 5.09 & 4.54 & 0.3242 & 0.0016 & - & - & - \\
        % LaDeCo~\cite{lin2024elements} & \textbf{8.08} & \underline{7.92} & \underline{8.00} & \textbf{7.82} & \textbf{6.98} & \textbf{0.0865} & \underline{0.0013} & - & - & - \\
        % \midrule
        % GPT-4o~\cite{hurst2024gpt} & \underline{7.48} & 7.43 & 7.59 & 7.13 & 6.41 & 0.1698 & \textbf{0.0012} & 0.8865 & 0.6788 & 0.9963 \\
        % \midrule
        % Ours & \textbf{8.08} & \textbf{7.99} & \textbf{8.11} & \underline{7.75} & \underline{6.96} & \underline{0.1020} & \underline{0.0013} & \textbf{0.8894} & \textbf{0.8426} & \textbf{0.9983} \\
        PosterLLaVa~\cite{yang2024posterllava} & - & - & - & - & - & \underline{0.0972} & \underline{0.0013} & \underline{0.8822} & \underline{0.8458} & - \\
        \midrule
        FlexDM~\cite{inoue2023document} & 5.34 & 5.29 & 5.41 & 5.09 & 4.54 & 0.3242 & 0.0016 & - & - & - \\
        LaDeCo~\cite{lin2024elements} & \underline{8.08} & \underline{7.92} & \underline{8.00} & \underline{7.82} & \underline{6.98} & \textbf{0.0865} & \underline{0.0013} & - & - & - \\
        \midrule
        GPT-4o~\cite{hurst2024gpt} & 7.41 & 7.31 & 7.54 & 7.09 & 6.37 & 0.1942 & \textbf{0.0011} & 0.8386 & 0.5544 & 0.9983 \\
        \midrule
        Ours & \textbf{8.25} & \textbf{8.10} & \textbf{8.24} & \textbf{7.87} & \textbf{7.10} & 0.0996 & \underline{0.0013} & \textbf{0.9150} & \textbf{0.8684} & \textbf{0.9991} \\
        % \midrule
        % GT & 8.35 & 8.21 & 8.30 & 8.01 & 7.26 & 0.0768 & 0.0015 \\
    \bottomrule
    \end{tabular}
    }
    \vspace{1mm}
    \caption{
    Quantitative comparison with baseline methods under the reconstruction setting.
    The best result for each metric is \textbf{bolded}, and the second-best result is \underline{underlined}.
    Note that for the baseline that does not consider the element content (i.e., PosterLLaVA), the LLaVA-OV scores are not applicable.
    For baselines that ignore element relationships (i.e., FlexDM and LaDeCo), Size Rel and Pos Rel are not applicable.
    For baselines that do not account for editing operations (i.e., PosterLLaVA, FlexDM, and LaDeCo), Op is not applicable.
    }
    \label{tab:quantitative_main}
\end{table*}

\subsection{Relation-Aware Design Reconstruction}
\label{subsec:self-supervise}

Since $\mathcal{G}$ is extracted from $\mathcal{A}_\text{in}$, we can replace $\mathcal{A}_\text{in}$ in $f: (\mathcal{C}, \mathcal{A}_\text{in}, \mathcal{O}) \mapsto \mathcal{A}_{\text{out}}$ (Section~\ref{subsec:overview}) with $\mathcal{G}$ to get an approximate mapping $\hat{f}: (\mathcal{C}, \mathcal{G}, \mathcal{O}) \mapsto \mathcal{A}$.
For simplicity, we remove the subscript of $\mathcal{A}_\text{out}$.
The new mapping makes it possible to train the layout editing model in a self-supervised manner, thus overcoming the data issue (Figure~\ref{fig:model}).
Since $\mathcal{C}$ and $\mathcal{A}$ in $\hat{f}$ can be directly extracted from a design $\mathcal{D}$, and $\mathcal{G}$ has already been introduced in Section~\ref{subsec:graph}, we will discuss how to obtain $\mathcal{O}$ below.

\noindent \textbf{Editing Operation.}
The first step is to sample an action from \{add, delete, move, resize\} and sample one element from $\mathcal{D}$ as the target element.
Second, based on the sampled action, we decide the parameters in the editing operation.
If the action is move, the parameters are the center-x and center-y coordinates of the target element.
If the action is resize, the parameters are the width and height of the target element.
If the action is add or delete, then no additional parameters are needed.
Putting the sampled action, the target element, and the parameters together, we synthesize a valid editing operation $\mathcal{O}$.
It is notable that we'll remove the edges related to the target element in $\mathcal{G}$, as the relationships between it and other nodes may change after editing.

\noindent \textbf{Sequence Representation.}
We devise the sequence representation for each component of $\hat{f}$.
(1) $\mathcal{C}$. 
We consider two element modalities in this work: image the text.
For image content, we use a vision encoder to convert it into image tokens.
Text content is naturally language tokens and does not require additional processing.
(2) $\mathcal{G}$.
We serialize it into a sequence by describing all edges in the format of ``node 1, relationship, node 2'', and concatenating these edges.
(3) $\mathcal{O}$.
We specify the editing action by its name (e.g., add, move).
The target element is indexed by an integer (e.g., element 5), and the required parameters are rounded to integers.
(4) $\mathcal{A}$.
Following previous work~\cite{lin2024elements}, we adopt JSON format to represent element attributes (see Figure ~\ref{fig:model}).
Image elements have four predicted attributes: center-x, center-y, width, and height.
For text elements, we predict three additional attributes: font size, text align, and angle.
We do not predict font type and color for text, because users typically do not want to modify them in the four actions.

\noindent\textbf{Model Architecture.}
We formulate design layout editing as a sequence-to-sequence transformation and fine-tune a multi-modal large language model (MLLM) to achieve it (Figure~\ref{fig:model}).
The MLLM consists of three key components: a vision encoder, a projector, and the LLM backbone.
The vision encoder is used to encode image content to produce image tokens.
The projector then projects them to match the dimensional requirement of the LLM backbone.
Other inputs, including the text content, the relation graph, and the editing operation, are directly fed into the LLM backbone, as they are already represented as discrete tokens.
The LLM backbone models the joint distribution of all inputs and generates the attributes auto-regressively.
\section{Experiments}
\label{label:exp}

\subsection{Setup}
\label{sec:setup}

\begin{figure*}
    \centering
    \includegraphics[width=\linewidth]{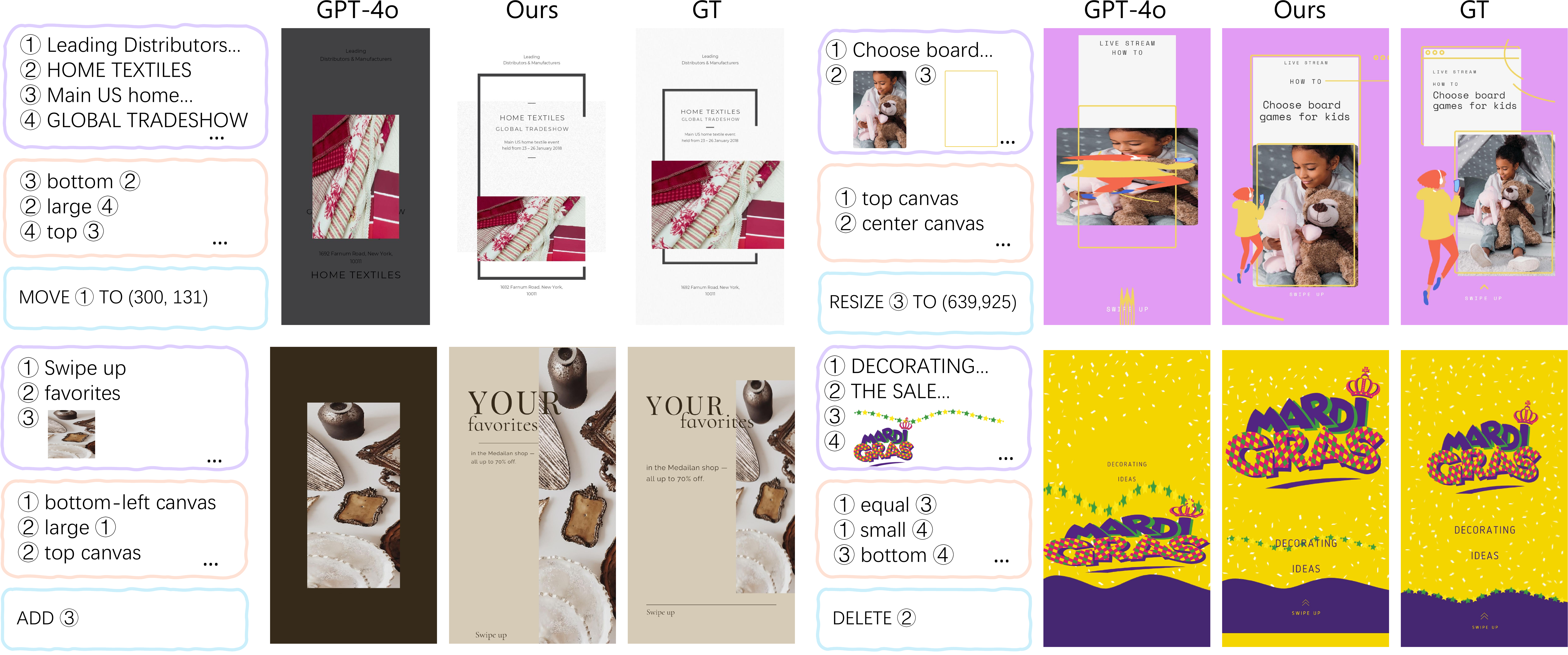}
    \caption{The reconstructed designs are visualized for qualitative comparison between our approach and GPT-4o. The ground truth (GT) designs are also shown. The three boxes on the left side of each case are the model inputs, i.e., \colorbox[rgb]{0.87,0.81,0.99}{element content}, \colorbox[rgb]{0.99,0.88,0.83}{element relationships}, and the \colorbox[rgb]{0.80,0.94,0.98}{editing operation}.
    Due to the large number of elements, we only visualize a subset of the content and relationships.}
    \label{fig:res_relation}
\end{figure*}

\noindent \textbf{Dataset.}
We used the publicly available dataset Crello~\cite{yamaguchi2021canvasvae} v4 in our experiments..
It contains 23,421 graphic designs, covering a variety of domains, such as cards, posters, and Instagram posts.
Each design has a set of metadata, including canvas size, element types, element positions, and so on.
For text elements, there are additional attributes like font, line height, color, etc.
The dataset is divided into training, validation, and test sets with 19,095, 1,951, 2,375 samples, respectively.
We follow LaDeCo~\cite{lin2024elements} to filter out the training samples with more than 25 elements.
To demonstrate the effectiveness of \approach{}, we introduce two test settings into the experiments.
(1) \textit{Reconstruction}. 
In this setting, the editing operation is extracted from the design in the same way as described in Section~\ref{subsec:self-supervise}.
It tests the reconstruction capability obtained \textit{directly} from RADR.
(2) \textit{Generalization}. 
This setting is more noteworthy and represents the goal of \approach{}.
It tests whether the model trained using RADR can generalize to the \textit{indirect} design layout editing task.
We simulate users to synthesize the editing operations for this setting.
Finally, both settings have a total of 2,375 (Crello test set size) * 2 (2 editing operations per test design) = 4,750 test samples.
Please refer to the supplementary materials for more details.

\noindent \textbf{Implementation Details.}
Following LaDeCo~\cite{lin2024elements}, we employ Llama-3.1-8B~\footnote{\href{https://huggingface.co/meta-llama/Llama-3.1-8B}{https://huggingface.co/meta-llama/Llama-3.1-8B}} as the backbone and CLIP ViT-L/14~\footnote{\href{https://huggingface.co/openai/clip-vit-large-patch14-336}{https://huggingface.co/openai/clip-vit-large-patch14-336}} as the vision encoder.
A two-layer MLP with GELU~\cite{hendrycks2016gaussian} activation functions serves as the projector, bridging the vision encoder and the backbone.
We apply LoRA~\cite{hu2021lora} fine-tuning to update the backbone and the projector (keeping the vision encoder frozen) on eight A100-40G GPUs, using AdamW~\cite{loshchilov2017decoupled} as the optimizer.
The global batch size is set to 64, the learning rate is 2e-4, and the LoRA rank is 32.

\noindent \textbf{Baselines.}
Since there is no prior work on this task, we choose GPT-4o~\cite{hurst2024gpt} as a baseline due to its powerful ability to handle multi-modal inputs.
The prompt for GPT-4o is included in the supplementary materials.
Additionally, we compare \approach{} with existing methods in two relevant categories:
(1) Design generation.
Since design layout editing ultimately produces a new design, we choose several advanced design generation methods, namely FlexDM~\cite{inoue2023document} and LaDeCo~\cite{lin2024elements}, as baselines to compare the output quality.
(2) Relationship-conditioned layout generation.
Given that \approach{} incorporates element relationships as generation conditions, we also compare it with PosterLLaVA~\cite{yang2024posterllava}, a specialized method for relationship-conditioned layout generation, assessing how well each approach adheres to the specified relationships.

\begin{figure*}[t]
  \begin{minipage}{0.49\textwidth}
    \flushleft
    \includegraphics[width=\linewidth]{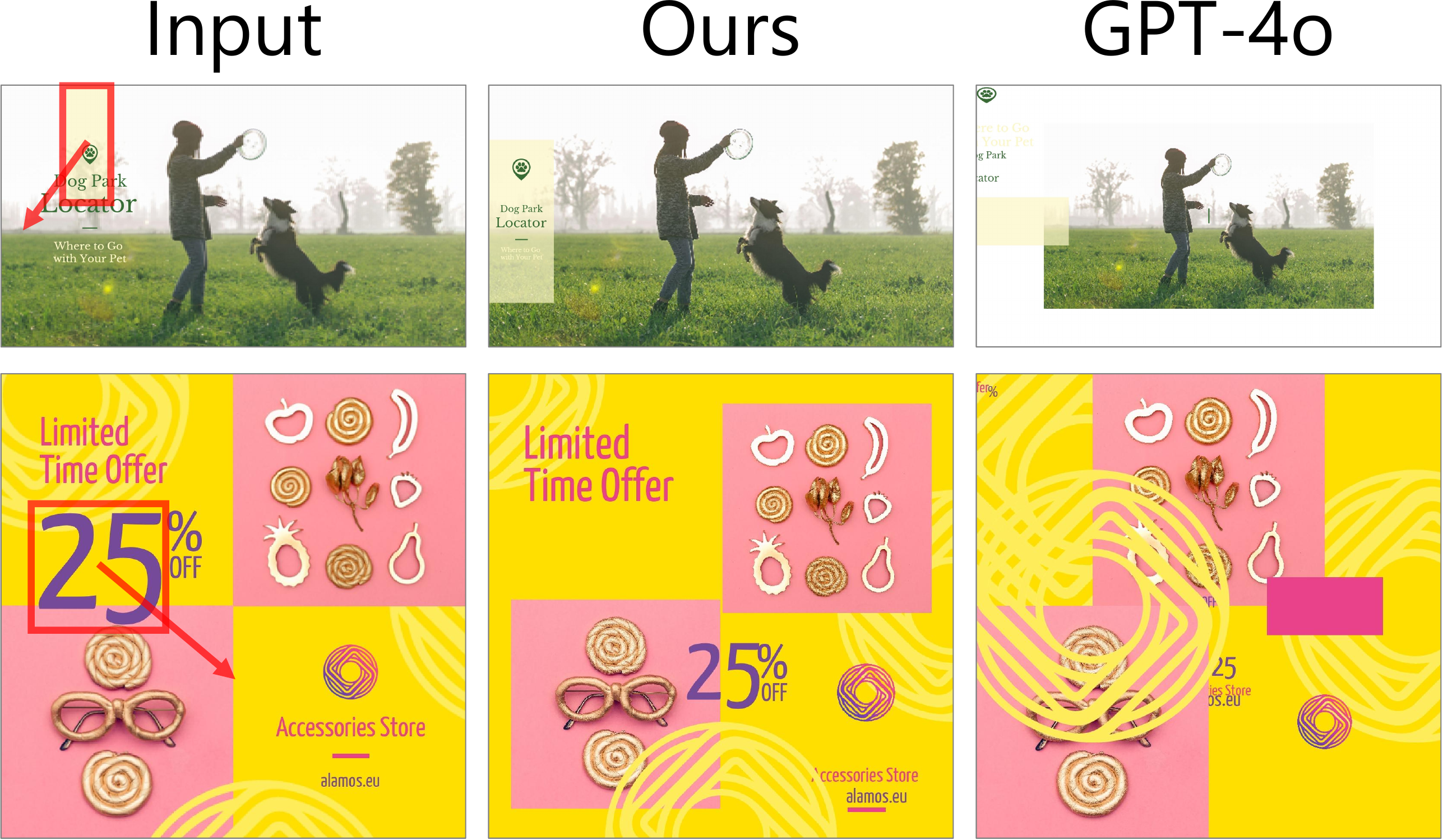}
    \caption{Qualitative comparison of the move action. The moved element is marked with a box. The arrow points to the target position.}
    \label{fig:res_move}
  \end{minipage}%
  \hfill
  \begin{minipage}{0.49\textwidth}
    \flushright
    \includegraphics[width=\linewidth]{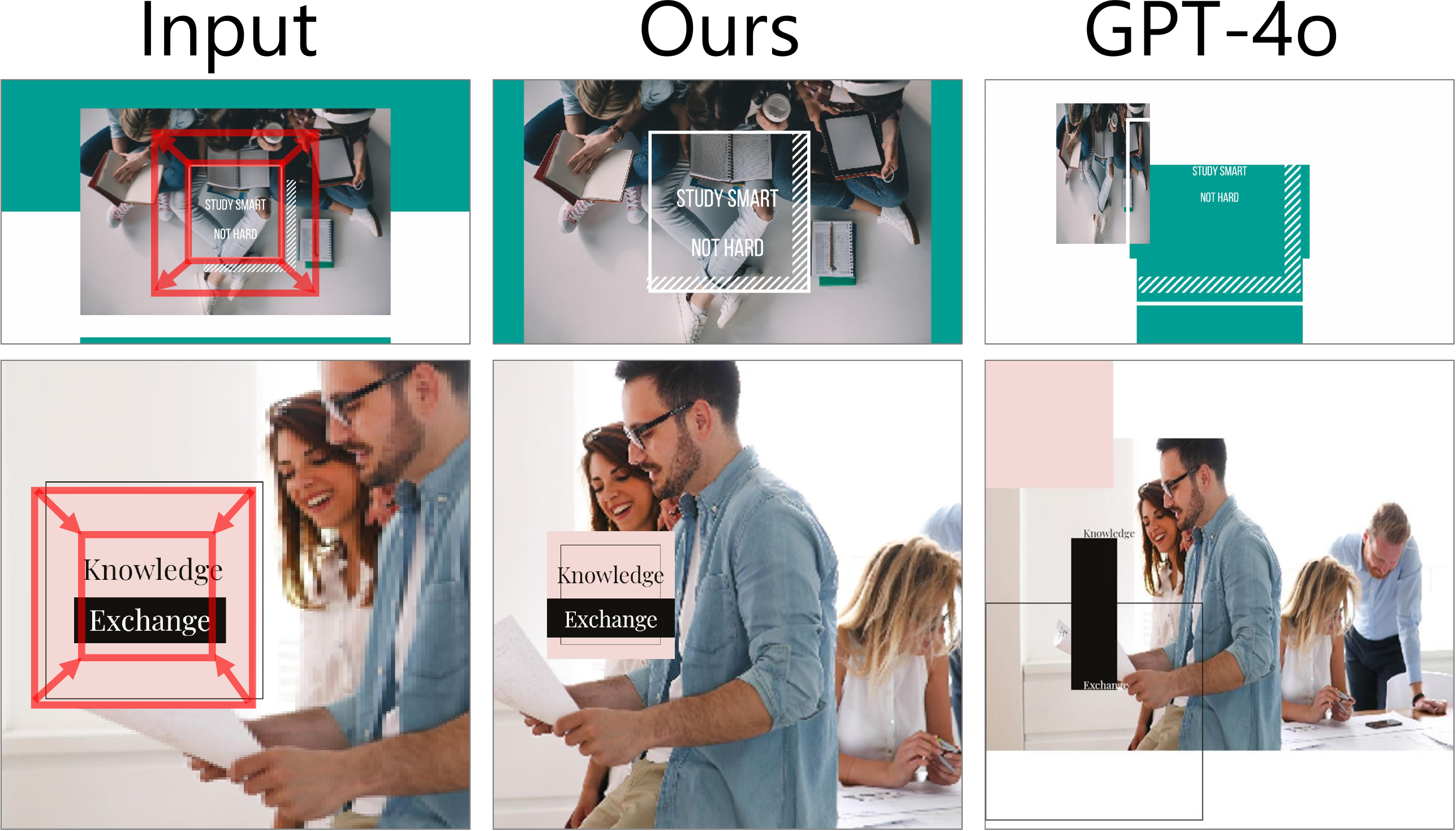}
    \caption{Qualitative comparison of the resize action. The resized element is marked with a box. The arrow points to the target size.}
    \label{fig:res_resize}
  \end{minipage}
\end{figure*}

\noindent \textbf{Metrics.}
Quantitative metrics involve three dimensions.
(1) Quality. 
We follow previous work~\cite{lin2024elements} to evaluate the design quality using LLaVA-OV scores~\cite{li2024llava}, element overlap (Ove), and alignment (Ali).
The LLaVA-OV scores assess quality from five perspectives: (i) design and layout, (ii) content relevance, (iii) typography and color, (iv) graphics and images, and (v) innovation and originality. 
(2) Layout structure preservation.
To assess whether the edited design preserves the layout structure of unedited elements, we present the relationship satisfaction rate (Rel).
Specifically, for each relationship in the relation graph, we check if it is satisfied (1) or not (0) in the output design.
The Rel metric is calculated as the ratio of the number of satisfied relationships to the total number of relationships. 
Based on the relationship category, we further break it down into size-related Rel (Size Rel) and position-related Rel (Pos Rel).
(3) Editing accuracy.
This dimension employs the editing operation satisfaction rate (Op).
Similar to the Rel metric, the Op value is 1 if the edited design meets the specified editing operation, and 0 otherwise.

\begin{figure*}[t]
  \begin{minipage}{0.49\textwidth}
    \flushleft
    \includegraphics[width=\linewidth]{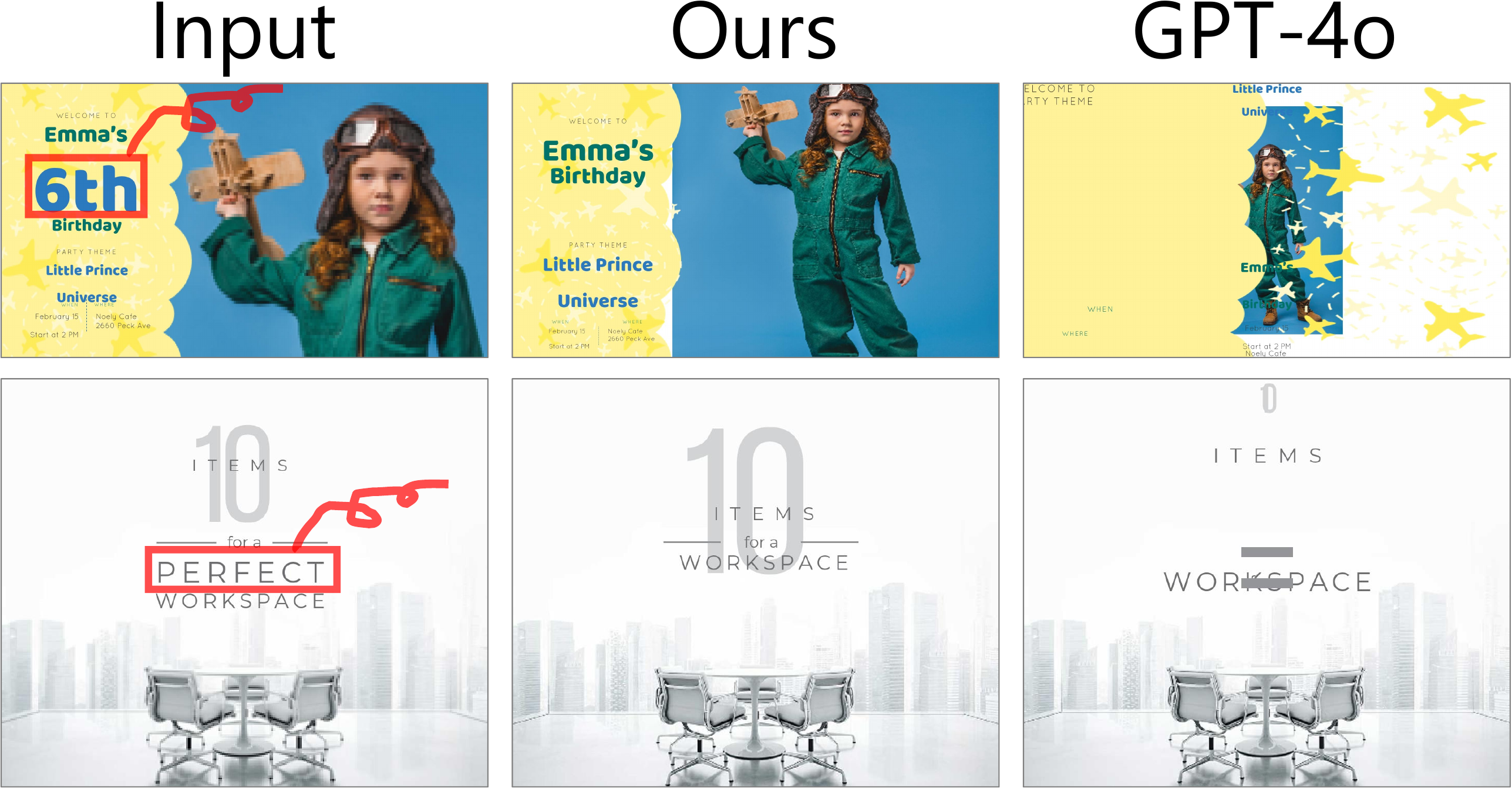}
    \caption{Qualitative comparison of the delete action. The deleted element is marked with a red box.}
    \label{fig:res_delete}
  \end{minipage}%
  \hfill
  \begin{minipage}{0.49\textwidth}
    \flushright
    \includegraphics[width=\linewidth]{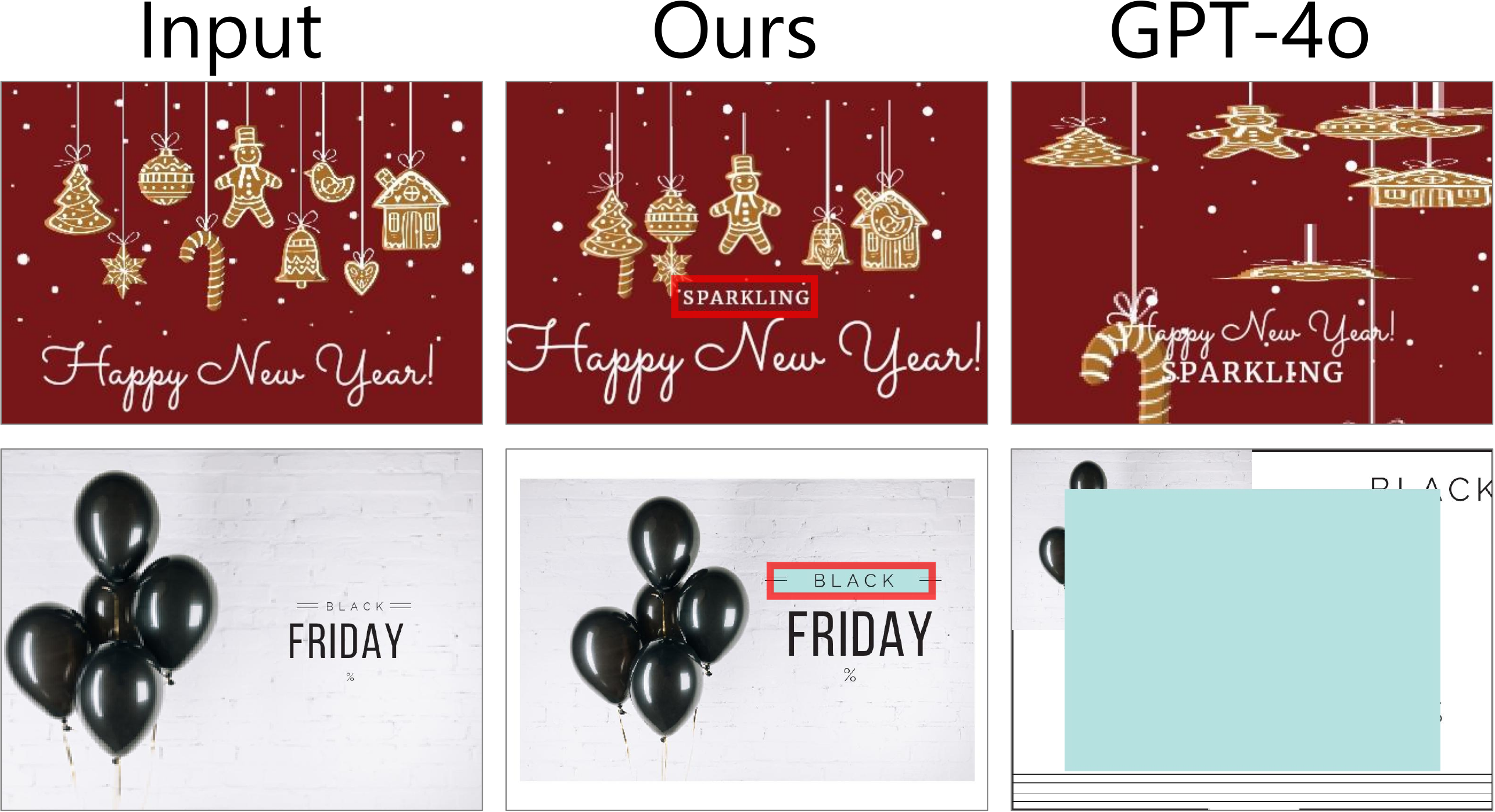}
    \caption{Qualitative comparison of the add action. The added element is marked with a red box.}
    \label{fig:res_add}
  \end{minipage}
\end{figure*}

\subsection{Quantitative Results}

The quantitative results of the reconstruction setting are shown in Table~\ref{tab:quantitative_main}.
\approach{} significantly outperforms GPT-4o in almost all metrics.
This demonstrates that our approach can perform high-quality and accurate design layout editing while effectively preserving the layout structure of unedited elements.
Additionally, \approach{} also showcases clear advantages over specialized baselines.
Specifically, \approach{} produces even better designs than the state-of-the-art design generation approach, i.e., LaDeco~\cite{lin2024elements}.
When compared to the relationship-conditioned layout generation model, \approach{} surpasses PosterLLaVA~\cite{yang2024posterllava} on two Rel metrics, which suggests its superiority in layout structure preservation.
Please see the supplementary materials for the quantitative results of the generalization setting.

\subsection{Qualitative Results}

Qualitative comparison of the reconstruction setting is presented in Figure~\ref{fig:res_relation}.
We also visualize the model inputs and the ground truth (GT) design for each case.
\approach{} outperforms the GPT-4o baseline in terms of design quality and relationship satisfaction, highlighting the strong reconstruction ability of our model.
For the generalization setting, the qualitative results are shown in Figures~\ref{fig:res_move},~\ref{fig:res_resize},~\ref{fig:res_delete}, and~\ref{fig:res_add}.
From the results, \approach{} exhibits the versatility in executing various editing actions.
Meanwhile, the results further showcase that \approach{} can produce high-quality edited designs while preserving the original layout structure in real editing scenarios.
In contrast, GPT-4o either suffers from severe quality issues (e.g., the second case of Figure~\ref{fig:res_add}, the first case of Figure~\ref{fig:res_delete}), or alters the layout structure (e.g., the first case of Figure~\ref{fig:res_resize}, the first case of Figure~\ref{fig:res_delete}).
Furthermore, we see that \approach{} can well adjust the unspecified elements to maintain visual harmony.
For example, in the second case of Figure~\ref{fig:res_move}, when the text "25" moves to a new position, the percent sign and the text "OFF" also move with it, retaining the semantics intact.
The qualitative comparison with more baselines can be found in the supplementary materials.

\subsection{Human Evaluation}

We also perform a human evaluation to compare \approach{} against GPT-4o.
To achieve this, we randomly sample 200 (original design, \approach{} edited design, GPT-4o edited design) triplets and evenly distribute them to four crowd workers for evaluation.
For each triplet, they are asked to answer two questions:
(1) Which of the two edited designs has better visual appeal?
(2) Which of the two edited designs better maintains the layout structure of the original design?
We conduct the evaluation under both settings, and show their percentage results in Table~\ref{tab:human_evaluation}.
From the results, we can see that \approach{} greatly surpasses GPT-4o in terms of layout structure preservation and generation quality, further proving the effectiveness of \approach{}.

\begin{table}
    \centering
    \begin{tabular}{cccc}
    \toprule
    Questions & \approach{} & Tie & GPT-4o \\
    \midrule
    \multicolumn{4}{c}{\textbf{\textit{Reconstruction Setting}}} \\
    (1) & 79.5 & 15.5 & 5.0 \\
    (2) & 86.0 & 13.0 & 1.0 \\
    \midrule
    \multicolumn{4}{c}{\textbf{\textit{Generalization Setting}}} \\
    (1) & 71.5 & 21.5 & 7.0 \\
    (2) & 86.0 & 10.5 & 3.5 \\
    \bottomrule
    \end{tabular}
    \vspace{1mm}
    \caption{Human evaluation results. The question (1) measures the output quality, and the question (2) reflects layout structure preservation. Under both settings, \approach{} significantly outperforms GPT-4o in both dimensions.}
    \vspace{-2mm}
    \label{tab:human_evaluation}
\end{table}

\begin{table*}[t]
    \centering
    \resizebox{0.9\textwidth}{!}{
    \begin{tabular}{lcccccccccc}
    \toprule
        \multirow{2}{*}{} & \multicolumn{5}{c}{LLaVA-OV Scores $\uparrow$} & \multirow{2}{*}{Ove $\downarrow$} & \multirow{2}{*}{Ali $\downarrow$} & \multirow{2}{*}{Size Rel $\uparrow$} & \multirow{2}{*}{Pos Rel $\uparrow$} & \multirow{2}{*}{Op $\uparrow$} \\
        % & Layout & Image & Typography & Content & Innovation & & & & & & & & \\
        & (i) & (ii) & (iii) & (iv) & (v) & & & & & \\
        \midrule
        % w/ random & 8.06 & 7.99 & 8.04 & 7.80 & 7.05 & 0.1007 & 0.0012 & 0.7884 & 0.4552 & 0.9993 \\
        % w/o relation graph & 8.15 & 8.05 & 8.18 & 7.78 & 7.02 & 0.0997 & 0.0013 & 0.7890 & 0.4610 & 0.9989 \\
        % w/ matrix & 8.05 & 7.96 & 8.04 & 7.76 & 7.05 & 0.1006 & 0.0013 & 0.8796 & 0.8332 & 0.9983 \\
        % Ours & 8.08 & 7.99 & 8.11 & 7.75 & 6.96 & 0.1020 & 0.0013 & 0.8894 & 0.8426 & 0.9983 \\
        % size 2 & 8.12 & 8.00 & 8.14 & 7.79 & 6.97 & 0.1059 & 0.0013 & 0.8505 & 0.7593 & & \\
        % size 8 & 8.11 & 7.99 & 8.16 & 7.81 & 7.00 & 0.1026 & 0.0014 & 0.8314 & 0.7166 & & \\
        % Ours & 8.11 & 7.98 & 8.15 & 7.80 & 7.01 & 0.1015 & 0.0014 & 0.8449 & 0.7339 & 0.9520 \\
        % \midrule
        % GT & 8.35 & 8.21 & 8.30 & 8.01 & 7.26 & 0.0768 & 0.0015 \\
        w/o RADR & 8.12 & 8.03 & 8.18 & 7.82 & 7.01 & 0.1007 & 0.0011 & 0.8008 & 0.3750 & 0.9978 \\
        w/o relation graph & 8.21 & 8.08 & 8.25 & 7.85 & 7.07 & 0.0980 & 0.0013 & 0.7943 & 0.3751 & 0.9987 \\
        w/ matrix & 8.17 & 8.07 & 8.20 & 7.84 & 7.04 & 0.0954 & 0.0012 & 0.8892 & 0.8529 & 0.9909 \\
        Ours & 8.25 & 8.10 & 8.24 & 7.87 & 7.10 & 0.0996 & 0.0013 & 0.9150 & 0.8684 & 0.9991 \\
    \bottomrule
    \end{tabular}
    }
    \vspace{1mm}
    \caption{Quantitative results of ablation studies. These experiments are all conducted and evaluated under the reconstruction setting.}
    \label{tab:ablation_study}
\end{table*}

\subsection{Ablation Studies}

\begin{figure*}[t]
  \begin{minipage}{0.48\textwidth}
    \flushright
    \includegraphics[width=\linewidth]{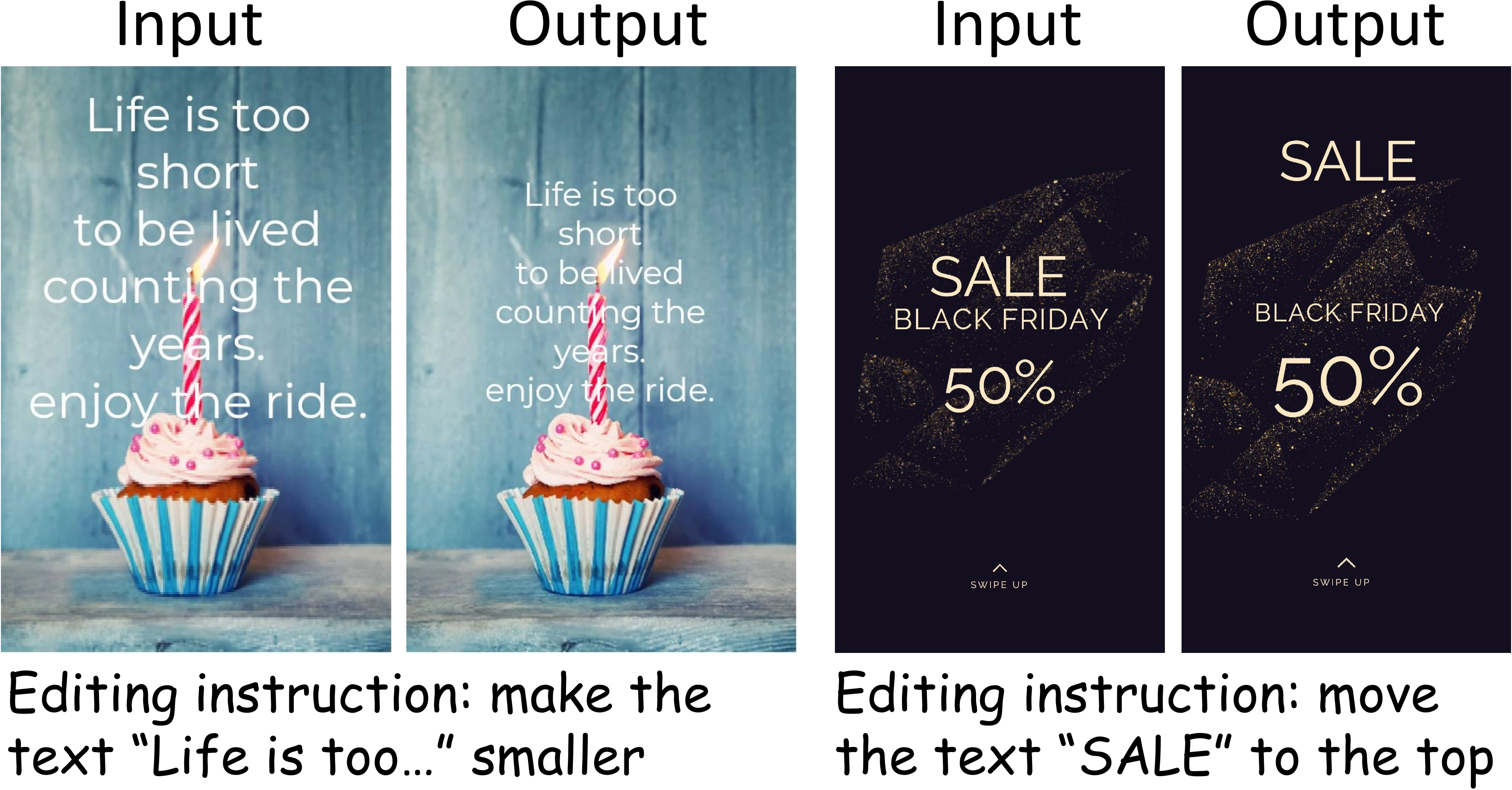}
    \caption{\approach{} can also perform design layout editing from textual instructions by integrating GPT-4o.}
    \label{fig:text_guided}
  \end{minipage}
  \hfill
  \begin{minipage}{0.45\textwidth}
    \flushleft
    \includegraphics[width=\linewidth]{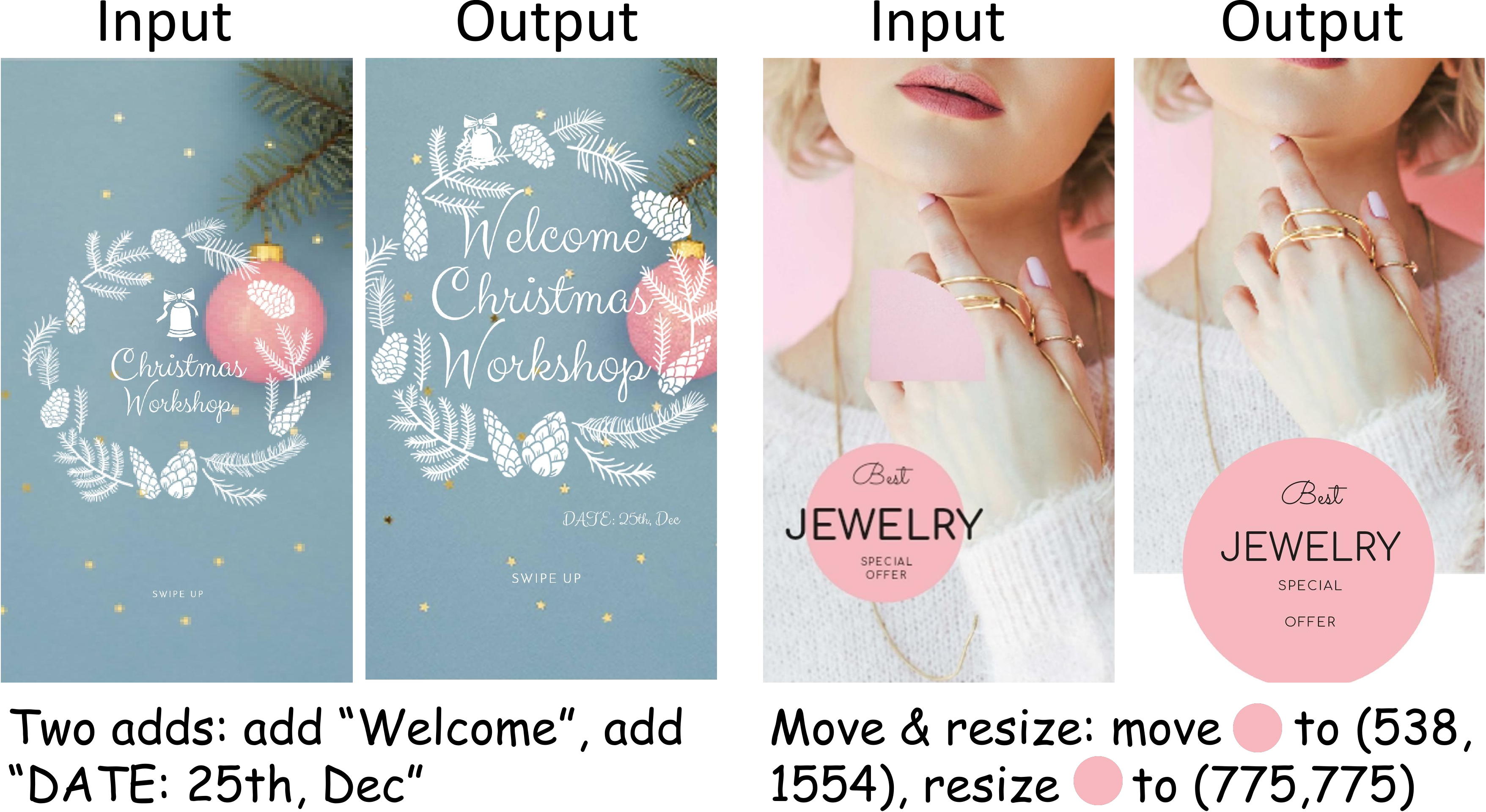}
    \caption{\approach{} can execute multiple editing operations at the same time, enhancing its practicality.}
    \label{fig:multi_op}
  \end{minipage}%
\end{figure*}

\noindent \textbf{Relation-Aware Design Reconstruction.}
Due to the lack of triplet data, relation-aware design reconstruction (RADR) is introduced into \approach{} to achieve design layout editing.
As a potential variation, we synthesize the original design by randomly sampling element attributes (e.g., uniformly sampling element width between 0 and the canvas width), thereby forming the (element content, random design, editing operation, good design) samples.
Its quantitative metrics are shown in Table~\ref{tab:ablation_study}, denoted as \textit{w/o RADR}.
The results demonstrate a significant drop in two Rel metrics, suggesting that this training strategy is inferior to our RADR in maintaining the layout structure.

\noindent \textbf{Relation Graph.}
In the work, we use the relation graph to represent the layout structure and guide the editing process for structure preservation.
To investigate its effect, we remove it, resulting in (element content, editing operation, edited design) samples for model training.
The results are shown in Table~\ref{tab:ablation_study} \textit{w/o relation graph}.
Similarly, the two Rel metrics are impaired, which indicates that the model cannot well preserve the layout structure without the relation graph.

\noindent \textbf{Graph Representation.}
In addition to the used sequence representation, we also employ an adjacency matrix to represent the relation graph (denoted as \textit{w/ matrix} in Table~\ref{tab:ablation_study}) and study the impact of different representations on model performance.
From the results, we see that the sequence representation achieves slightly better Rel metrics than the matrix representation, which demonstrates that the compact sequence format is more effective in representing the relation graph.
Furthermore, the close performance of both representations highlights the robustness of \approach{}.

\subsection{Applications}

\noindent \textbf{Language-Guided Design Layout Editing.}
For natural language editing instructions, we leverage GPT-4o~\cite{hurst2024gpt} to translate them into corresponding editing operations.
The original design is also used as input during the translation process.
\approach{} then executes the translated operation, achieving language-guided design layout editing.
Figure~\ref{fig:text_guided} shows some results.
This significantly improves the usability of \approach{}, as expressing editing intentions in natural language is more intuitive and user-friendly.

\noindent \textbf{Composite Design Layout Editing.}
While \approach{} is trained on individual operations, it can be readily extended to support composite operations at test time (see Figure~\ref{fig:multi_op}). 
Such flexible combination enables \approach{} to handle complex editing needs that are not seen during training.

\section{Conclusion}

In this work, we propose \approach{} to address the underexplored design layout editing task.
By introducing the relation graph for structural representation and a novel relation-aware design reconstruction strategy, we achieve versatile and structure-preserving layout design in the absence of triplet data.
Extensive quantitative and qualitative experiments, user studies, alongside the supported applications, demonstrate the model's effectiveness.
Despite these advancements, limitations remain. 
Specifically, when the element number in the input design grows, it becomes more and more challenging to preserve the overall layout structure and ensure the design quality while performing editing.
The typical failure cases include: (1) the violations of specified relationships, and (2) quality issues such as unintended overlaps.
Future work will extend the system's capabilities to include font, color, and content editing, as well as develop a more sophisticated language-guided interface. 
We believe \approach{} serves as a foundational step toward more democratized and intelligent design workflows.

\newpage

% In the unusual situation where you want a paper to appear in the
% references without citing it in the main text, use \nocite

\newpage

\bibliography{example_paper}
\bibliographystyle{icml2026}

%%%%%%%%%%%%%%%%%%%%%%%%%%%%%%%%%%%%%%%%%%%%%%%%%%%%%%%%%%%%%%%%%%%%%%%%%%%%%%%
%%%%%%%%%%%%%%%%%%%%%%%%%%%%%%%%%%%%%%%%%%%%%%%%%%%%%%%%%%%%%%%%%%%%%%%%%%%%%%%
% APPENDIX
%%%%%%%%%%%%%%%%%%%%%%%%%%%%%%%%%%%%%%%%%%%%%%%%%%%%%%%%%%%%%%%%%%%%%%%%%%%%%%%
%%%%%%%%%%%%%%%%%%%%%%%%%%%%%%%%%%%%%%%%%%%%%%%%%%%%%%%%%%%%%%%%%%%%%%%%%%%%%%%
\newpage
\appendix
\onecolumn

\section{More Implementation Details}

\subsection{Data Example}
\label{sec:supp_data_example}

We present an example to demonstrate the model input and output in \approach{}.
For the image element (e.g., element 0, 1, 2), our model predicts its center-x, center-y, width, and height.
For the text element (e.g., element 3), the model also predicts its angle, font size, and text align.

\begin{coloredtextbox1}[A data example of \approach{}.]{orange}
\textbf{Input}: 

CANVAS SIZE: \{``width": 940, ``height": 788\}

ELEMENT CONTENT: [\{``index": 0, ``content": ``\textless image\textgreater"\}, \{``index": 1, ``content": ``\textless image\textgreater"\}, \{``index": 2, ``content": ``\textless image\textgreater"\}, \{``index": 3, ``content": ``STOP DREAMING START DOING"\}]

SIZE RELATIONSHIP: [``element 0 large element 1", ``element 2 small element 1", ...]

POSITION RELATIONSHIP: [``element 1 center element 0", ``element 2 top canvas", ...]

EDITING OPERATION: move element 3 to \{``x": 583, ``y": 394\}

\vspace{1mm}

\textbf{Output}: 
\begin{Verbatim}[frame=single]
[
{"index": 0, "x": 470, "y": 394, "width": 940, "height": 806}, 
{"index": 1, "x": 470, "y": 394, "width": 873, "height": 721}, 
{"index": 2, "x": 598, "y": 147, "width": 443, "height": 418}, 
{"index": 3, "x": 583, "y": 394, "width": 441, "height": 336, 
"angle": 0, "font_size": 70, "text_align": "left"}
]
\end{Verbatim}

\end{coloredtextbox1}

\subsection{GPT-4o Prompt}

We also present the prompt used in the GPT-4o baseline.
The \textit{model input} part is the same as that in Section~\ref{sec:supp_data_example} for fair comparison.

\begin{coloredtextbox1}[The prompt for the GPT-4o baseline.]{gray}
You are an autonomous AI assistant who aids designers to perform design layout editing by predicting element attributes based on the canvas size, element content, element size relationships, element position relationships, and an editing operation.

\vspace{1mm}

Please ensure that the specified element relationships are satisfied in the output design. For example, ``element 1 large element 2'' denotes that element 1 should be larger than element 2, ``element 1 top-left canvas'' represents that element 1 should be placed at the top-left of the canvas.
Besides, the editing operation also should be taken into account.

\vspace{1mm}

For the image element, you need to predict the center-x, center-y, width, and height attributes. For the text element, you should additionally predict its angle, font size, and text align. Only respond in JSON format, no other information. Here is an example output:

\textit{(an example output)} \ \ \ \ NOTE: Its format is the same as that in Section~\ref{sec:supp_data_example}.

\vspace{1mm}

\textbf{Input}: 

\textit{(model input)} \ \ \ \ NOTE: Its format is the same as that in Section~\ref{sec:supp_data_example}.

\vspace{1mm}

\textbf{Output}: 

\textit{(model output)}

\end{coloredtextbox1}

\subsection{Model Testing}

To test the performance of \approach{}, we curate two evaluation settings on the test set of Crello: reconstruction and generalization, as described below.
\paragraph{Reconstruction.}
Since \approach{} exploits a relation-aware design reconstruction (RADR) strategy to achieve design layout editing without triplet data, it is necessary to evaluate its reconstruction ability.
In this setting, the test samples are obtained in the same way as described in Section~\ref{subsec:self-supervise}.
This is to say, given an input design, we use its element content, the relation graph, and an editing operation extracted from it as model input, and see how similar the output design is to this design (i.e., ground truth) in layout structure (measured by Size Rel, Pos Rel).
Besides, we also evaluate the quality of the output design.
For greater clarity, we elaborate on the extracted editing operation of each editing action below:

\begin{itemize}[leftmargin=10pt]
  \item Move: The target element is selected by sampling one element from the input design. The parameters are the real center-x and center-y coordinates of the target element.
  \item Resize: The target element is selected by sampling one element from the input design. The parameters are the real width and height of the target element.
  \item Add: The target element is selected by sampling one element from the input design. There are no parameters in this action.
  \item Delete: Since deleting an element from the given design will change the number of elements, instead, we sample an element from the validation set of Crello as the target element. There are no parameters in this action.
\end{itemize}

\paragraph{Generalization.}

In this setting, we test the generalization ability of \approach{} by simulating user-specified editing operations.
To achieve this, we sample the parameters for the editing operation instead of extracting them from the input design.
Hence, there is no ground truth under this setting.
Specifically,

\begin{itemize}[leftmargin=10pt]
  \item Move: The target element is selected by sampling one element from the input design. Its new center-x coordinate is uniformly sampled from the range [0, canvas width]. Similarly, the new center-y coordinate is uniformly sampled from [0, canvas height].
  \item Resize: The target element is selected by sampling one element from the input design. The resizing factor is uniformly sampled from the range [0.5, 2.0]. The element's new width in the editing operation is computed by multiplying its original width by the resizing factor. Similarly, its new height is obtained by multiplying its original height by the same factor.
  \item Add: The target element is sampled from the validation set of Crello. There are no parameters in this action.
  \item Delete: The target element is selected by sampling one element from the input design. There are no parameters in this action.
\end{itemize}

\section{More Experimental Results}

\subsection{Quantitative Results of the Generalization Setting}

We have included the quantitative results of the reconstruction setting in the main body.
Here, we present the quantitative results of the generalization setting in Table~\ref{tab:customized}.
\approach{} significantly outperforms GPT-4o on almost all metrics.

\begin{table*}[ht]
    \centering
    \begin{small}
    \begin{tabular}{lcccccccccc}
    \toprule
        \multirow{2}{*}{Methods} & \multicolumn{5}{c}{LLaVA-OV Scores $\uparrow$} & \multirow{2}{*}{Ove $\downarrow$} & \multirow{2}{*}{Ali $\downarrow$} & \multirow{2}{*}{Size Rel $\uparrow$} & \multirow{2}{*}{Pos Rel $\uparrow$} & \multirow{2}{*}{Op $\uparrow$} \\
        % & Layout & Image & Typography & Content & Innovation & & & & & & & & \\
        & (i) & (ii) & (iii) & (iv) & (v) & & & & & \\
        \midrule
        GPT-4o & 7.23 & 7.20 & 7.39 & 6.92 & 6.21 & 0.1789 & 0.0013 & 0.8447 & 0.5444 & 0.9979 \\
        Ours & 8.19 & 8.14 & 8.24 & 7.88 & 7.11 & 0.0949 & 0.0009 & 0.9475 & 0.9157 & 0.9960 \\
    \bottomrule
    \end{tabular}
    \end{small}
    \caption{Quantitative comparison with GPT-4o under the generalization setting.}
    \label{tab:customized}
\end{table*}

\subsection{Extension to Novel Actions}

To evaluate the versatility and generalization of \approach{}, we extend its application beyond the four actions to more complex, multi-action scenarios. Specifically, we define three novel composite actions by combining existing ones:

\begin{itemize}[leftmargin=10pt]
  \item Replace: A combination of delete and add.
  \item Double Add: Simultaneous addition of two elements.
  \item MoveResize: Concurrent adjustment of an element's position and size.
\end{itemize}

We curate a test set comprising 200 samples for each composite action. Input designs are randomly sampled from the Crello test set.
Notably, the combination of actions is quite flexible and is not limited to these three. 
For example, ``add + move" means adding an element to a specific position, ``add + resize" means adding an element with a specific size.

A key strength of \approach{} is its ability to handle these composite actions at inference time without requiring additional training or architectural modifications. 
As shown in Table~\ref{tab:novel_apps}, our model achieves competitive performance across all three novel applications without requiring action-specific training, maintaining high editing quality and relationship preservation.
The quantitative results demonstrate that \approach{} exhibits robust generalization to unseen action combinations. 

In conclusion, \approach{} is the first design layout editing approach that supports a wide range of editing actions. 
Beyond the four actions discussed in the paper, the combination of multiple actions also extends to other novel and unseen actions. 
The experimental results highlight the key strengths of our approach: its strong generalization capability, broad applicability, support for diverse editing actions, and remarkable flexibility.

\begin{table*}[t]
    \centering
    \resizebox{0.9\textwidth}{!}{
    \begin{tabular}{lcccccccccc}
    \toprule
        \multirow{2}{*}{} & \multicolumn{5}{c}{LLaVA-OV Scores $\uparrow$} & \multirow{2}{*}{Ove $\downarrow$} & \multirow{2}{*}{Ali $\downarrow$} & \multirow{2}{*}{Size Rel $\uparrow$} & \multirow{2}{*}{Pos Rel $\uparrow$} & \multirow{2}{*}{Op $\uparrow$} \\
        % & Layout & Image & Typography & Content & Innovation & & & & & & & & \\
        & (i) & (ii) & (iii) & (iv) & (v) & & & & & \\
        \midrule
        Replace         & 8.13 & 8.05 & 8.18 & 7.85 & 7.04 & 0.0978 & 0.0012 & 0.8840 & 0.8410 & 1.0000 \\
Double Add      & 8.10 & 8.03 & 8.13 & 7.78 & 7.04 & 0.1126 & 0.0013 & 0.8911 & 0.8375 & 1.0000 \\
MoveResize      & 8.20 & 8.06 & 8.22 & 7.85 & 7.11 & 0.0956 & 0.0009 & 0.8744 & 0.8467 & 0.9850 \\
\midrule
{Ours (Gen.)} & {8.19} & {8.14} & {8.24} & {7.88} & {7.11} & {0.0949} & {0.0009} & {0.9475} & {0.9157} & 0.9960 \\
\bottomrule
    \end{tabular}
    }
    \vspace{1mm}
    \caption{Quantitative results of \approach{} on novel editing actions.}
    \label{tab:novel_apps}
\end{table*}

\subsection{More Qualitative Comparison of the Generalization Setting}

\begin{figure}
    \centering
    \includegraphics[width=\linewidth]{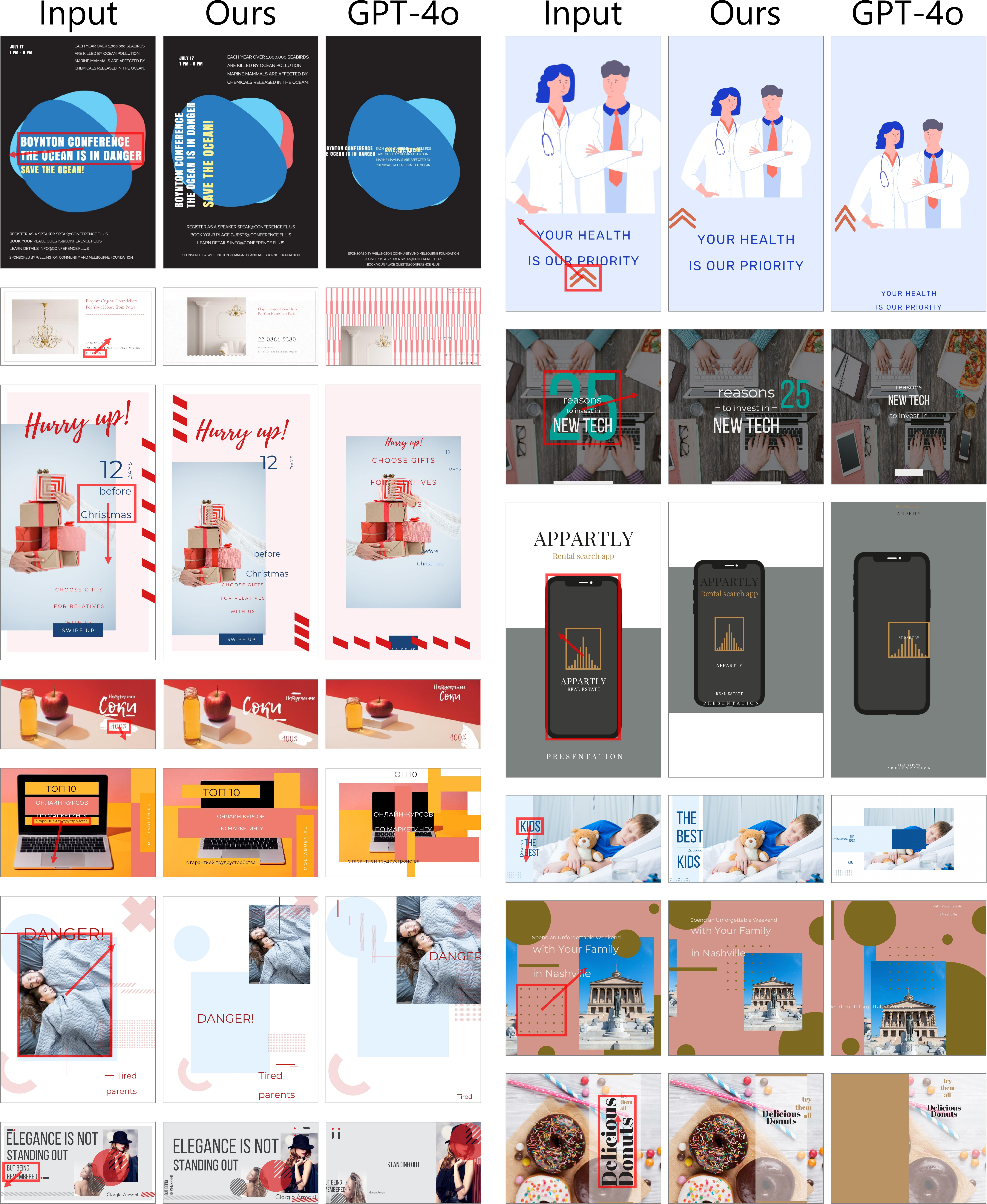}
    \caption{More qualitative comparison of the move action. The edited element and the target position are marked in the input design.}
    \label{fig:move_supp}
\end{figure}

\begin{figure}
    \centering
    \includegraphics[width=\linewidth]{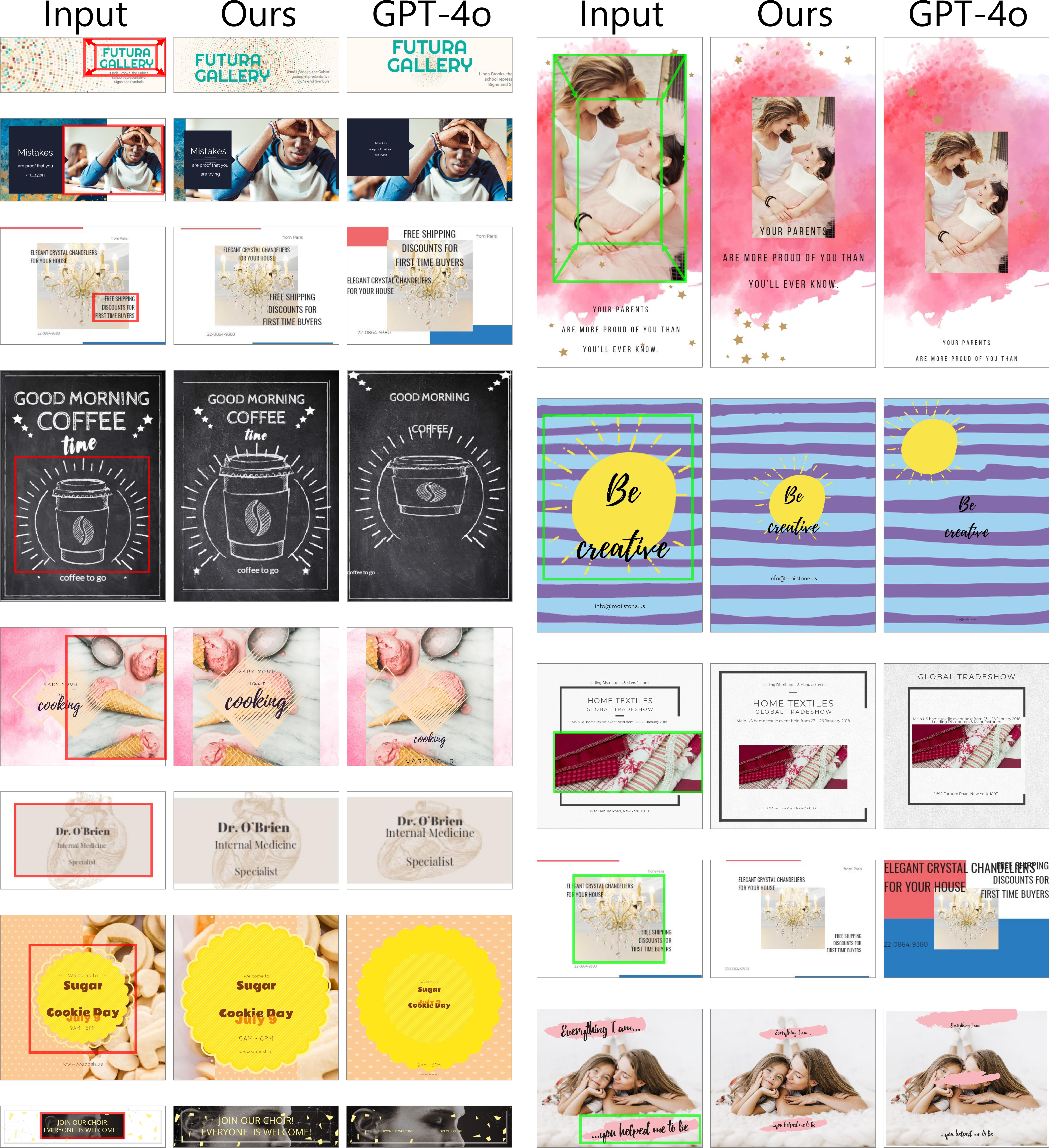}
    \caption{More qualitative comparison of the resize action. The edited element is marked in the input design, with \textit{red} indicating the zoom-in resize action (left side) and \textit{green} indicating the zoom-out action (right side).}
    \label{fig:resize_supp}
\end{figure}

\begin{figure}
    \centering
    \includegraphics[width=\linewidth]{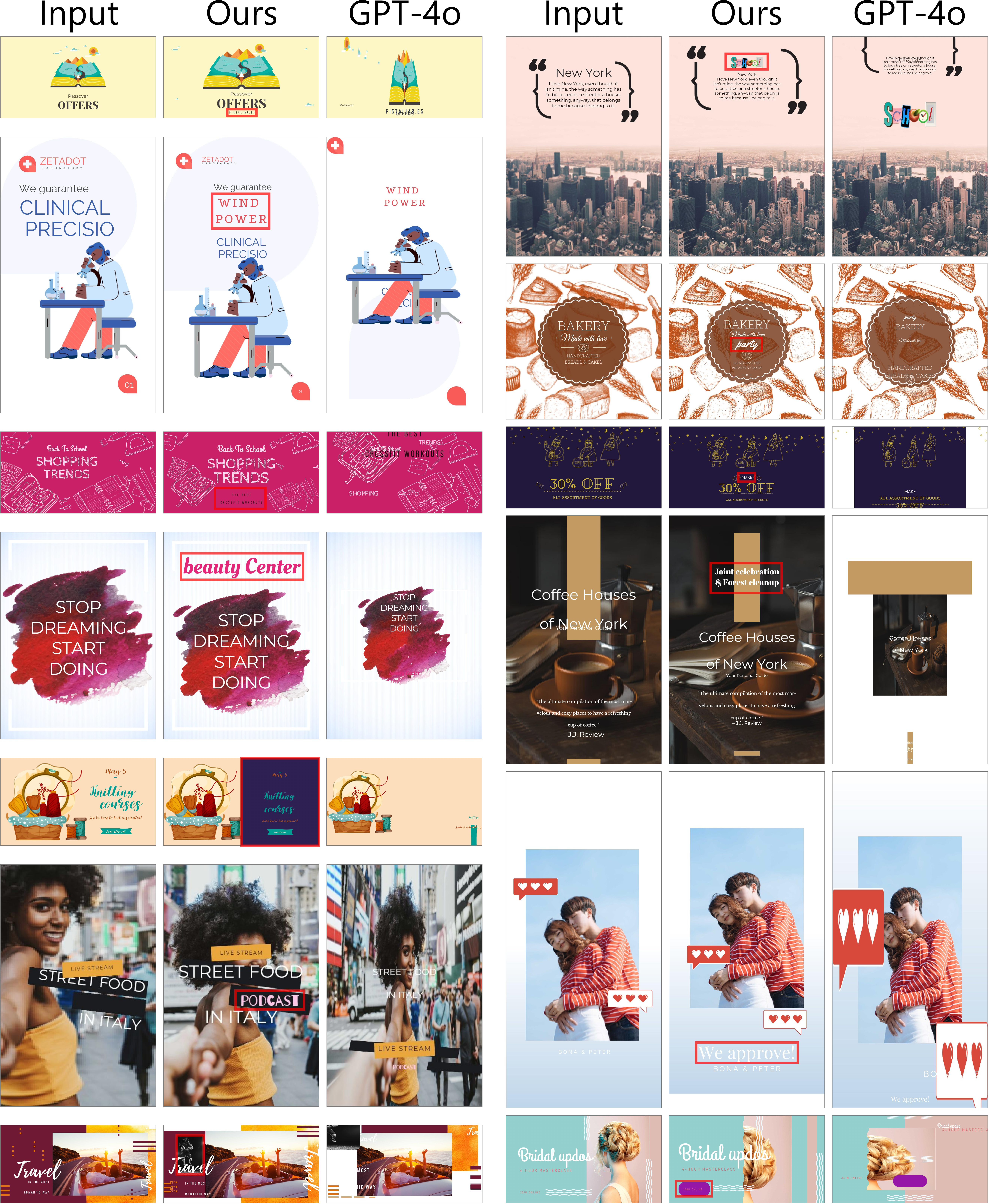}
    \caption{More qualitative comparison of the add action. The added element is marked in the output design of our approach.}
    \label{fig:add_supp}
\end{figure}

\begin{figure}
    \centering
    \includegraphics[width=\linewidth]{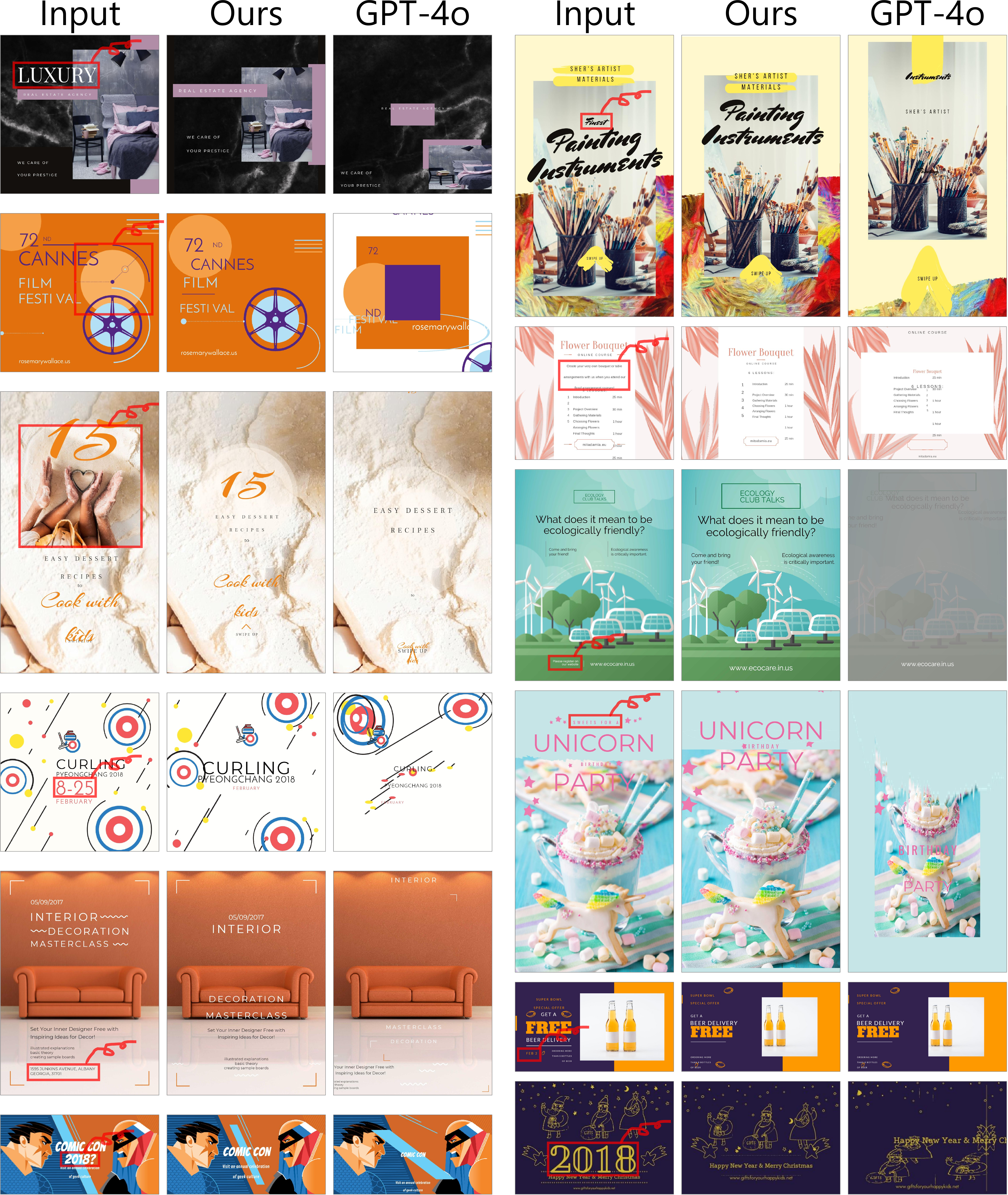}
    \caption{More qualitative comparison of the delete action. The deleted element is marked in the input design.}
    \label{fig:delete_supp}
\end{figure}

We show more qualitative results of the generalization setting in Figures~\ref{fig:move_supp}, \ref{fig:resize_supp}, \ref{fig:add_supp}, and \ref{fig:delete_supp} to demonstrate the effectiveness of \approach{}.
The results show that 
(1) \approach{} is a versatile approach that can accurately perform different kinds of editing actions using a single model, 
(2) \approach{} preserves the layout structure of the input design well in the edited design,
(3) \approach{} can produce high-quality edited designs.

\subsection{Quality Comparison}

\begin{figure}
    \centering
    \includegraphics[width=\linewidth]{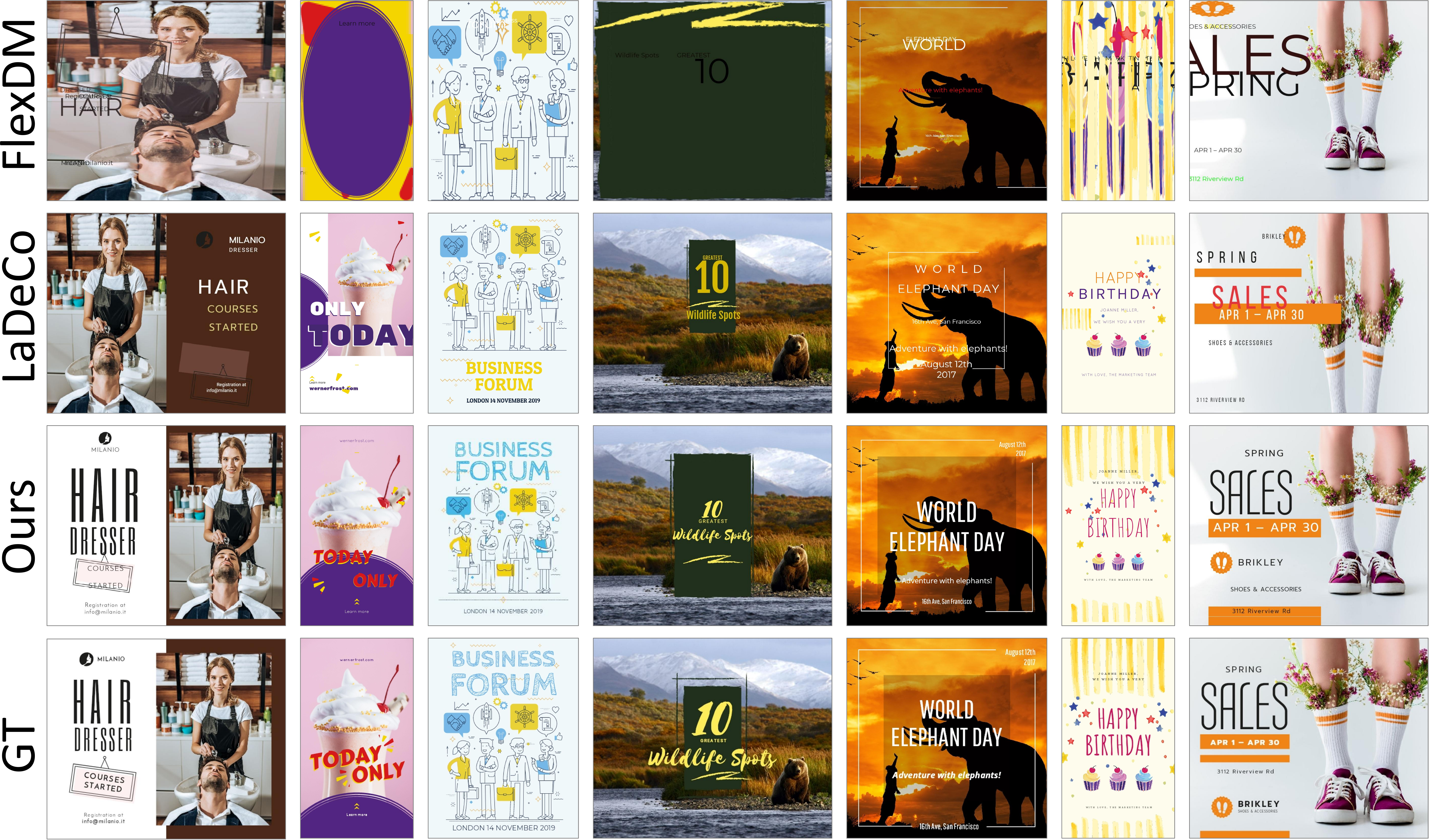}
    \caption{Qualitative comparison with the existing design generation approaches under the reconstruction setting. The ground truth (GT) designs are also visualized.}
    \label{fig:design_quality_supp}
\end{figure}

We qualitatively compare \approach{} with existing design generation models, including LaDeCo and FlexDM, as shown in Figure~\ref{fig:design_quality_supp}.
Our method produces design outputs that are on par with the state-of-the-art model LaDeCo in terms of visual quality.
Moreover, the comparison with ground-truth (GT) designs highlights the strong reconstruction capability of \approach{}.

\section{Efficiency Analysis}

The average inference time of our approach is 14.8 seconds per example. 
This is slower than FlexDM (0.3 s/sample), a non-autoregressive generation model, and PosterLLaVa (12.0 s/sample), which does not encode all image content in the prompt, resulting in a shorter input sequence. 
On the other hand, our approach is more efficient than LaDeCo (17.5 s/sample), which involves an intermediate rendering process, and GPT-4o, which is constrained by API call rate limitations.

%%%%%%%%%%%%%%%%%%%%%%%%%%%%%%%%%%%%%%%%%%%%%%%%%%%%%%%%%%%%%%%%%%%%%%%%%%%%%%%
%%%%%%%%%%%%%%%%%%%%%%%%%%%%%%%%%%%%%%%%%%%%%%%%%%%%%%%%%%%%%%%%%%%%%%%%%%%%%%%

\end{document}